\documentclass{article}

\usepackage{microtype}
\usepackage{graphicx}
\usepackage{subcaption}
\usepackage{float}
\usepackage{booktabs} 
\usepackage{fancyhdr,graphicx,amsmath,amssymb}
\usepackage[ruled,vlined]{algorithm2e}
\include{pythonlisting}

\usepackage{hyperref}

\usepackage[accepted]{icml2023}

\usepackage{amsmath}
\usepackage{amssymb}
\usepackage{mathtools}
\usepackage{amsthm}
\usepackage[capitalize,noabbrev]{cleveref}

\def\tcb{\textcolor{blue}}
\usepackage{verbatim}
\usepackage{cuted}
\usepackage{mathtools, bm}
\usepackage{amssymb,amsfonts,latexsym,color}
\usepackage{comment}
\usepackage{wrapfig}

\theoremstyle{plain}

\theoremstyle{definition}

\theoremstyle{remark}

\usepackage[textsize=tiny]{todonotes}

\icmltitlerunning{LESSON: Learning to Integrate Exploration Strategies for Reinforcement Learning via an Option Framework}

\begin{document}

\twocolumn[
\icmltitle{LESSON: Learning to Integrate Exploration Strategies for Reinforcement Learning via an Option Framework}

\icmlsetsymbol{equal}{*}

\begin{icmlauthorlist}
\icmlauthor{Woojun Kim}{equal,KAIST}
\icmlauthor{Jeonghye Kim}{equal,KAIST}
\icmlauthor{Youngchul Sung}{KAIST}
\end{icmlauthorlist}

\icmlaffiliation{KAIST}{School of Electrical Engineering, KAIST, Daejeon 34141, Republic of Korea}

\icmlcorrespondingauthor{Youngchul Sung}{ycsung@kaist.ac.kr}

\icmlkeywords{Exploration, Exploitation-Exploration Trade-off, Option-Critic}

\vskip 0.3in
]



\printAffiliationsAndNotice{\icmlEqualContribution} 

\begin{abstract}
In this paper, a unified framework for exploration in reinforcement learning (RL) is proposed based on an option-critic model. The proposed framework learns to integrate a set of diverse exploration strategies so that the agent can adaptively select the most effective exploration strategy over time to realize a relevant exploration-exploitation trade-off for each given task. The effectiveness of the proposed exploration framework is demonstrated by various experiments in the MiniGrid and Atari environments. 
\end{abstract}

\section{Introduction}\label{sec:intro}

RL is a powerful framework to obtain an  optimal policy that maximizes the expected return by learning from experiences. However, the convergence to an optimal policy by model-free RL requires that all state-action pairs should be visited infinitely often  \cite{sutton2018reinforcement}, but this is impractical in real-world situations with limited time and resources. Therefore, efficient exploration  has been one of the core research topics of RL throughout its history and many sophisticated methods have been proposed recently for efficient exploration, e.g., temporally-extended exploration \cite{osband2016deep, dabney2020temporally, yu2021taac},    intrinsic motivation-based exploration \cite{bellemare2016unifying, achiam2017surprise, burda2018exploration}, 
maximum entropy RL \cite{haarnojaSQL,haarnoja2018soft,HanSungDAC,HanSungMaxMin,KimSungMMI,KimSungADER}, parallel search \cite{JaderbergPBT,JungSungP3S}. However,   there is no single exploration method found yet  that is shown to be  universally effective across all tasks.  
For example,  intrinsic motivation-based exploration, which augments  extrinsic reward with additional exploration-oriented intrinsic reward,  is shown to be effective for hard exploration tasks but to have negative effects on some environments with dense rewards. Temporally-extended exploration, which encourages the temporal persistence of exploration, tends to outperform simple exploration strategies such as $\epsilon$-greedy and  Gaussian noise injection, but struggles to solve hard exploration tasks. In some cases, $\epsilon$-greedy even performs better than the aforementioned methods. Thus, for the best performance, one needs to  try various exploration methods and select the best one for each given task.  However, this is  a time-consuming and difficult task.   
In addition to the variability of  good exploration strategy across  tasks, the required exploration strategy can even vary over time during the training period  within a given task. Thus, one selected exploration strategy may not be optimal throughout the whole training period for a given task.

In this paper, we address  such variability of good exploration strategy   and propose a  unified  exploration framework named LESSON, aiming at universality  across tasks and learning phases.  In our framework,  the  agent learns to integrate a set of diverse exploration strategies  so that it can  automatically select  the most effective  exploration strategy for each phase of learning  for each given task from the context of exploration-exploitation trade-off.  
To devise such a unification framework for multiple exploration strategies, we adopt an {\em option-critic model} \cite{bacon2017option}. However, simple application of an  option-critic model to exploration strategies does not yield the desired unification. We circumvent this difficulty by employing  off-policy learning and judiciously designing the overall off-policy structure with    objective functions and action value functions suitable to our off-policy exploration-exploitation trade-off.  We show that LESSON can achieve significant performance improvement over existing exploration methods.   To the best of our knowledge, LESSON is the first unified framework  that can learn to integrate  multiple exploration strategies for adaptive exploration strategy selection targeting relevant exploration-exploitation trade-off  over the learning phase.

\section{Background and Related Works}\label{sec:background}

We consider a Markov decision process (MDP) defined as a tuple $<\mathcal{S}, \mathcal{A}, \mathcal{P}, R, \gamma>$, where $\mathcal{S}$ is the state space,  $\mathcal{A}$ is the action space,  $\mathcal{P}:\mathcal{S}\times \mathcal{A}\times \mathcal{S} \rightarrow [0,1]$ is the transition probability,  $R:\mathcal{S} \times \mathcal{A} \rightarrow {\mathbb{R}}$ is the reward function, and  $\gamma \in [0,1)$ is the discount factor. At each time step $t$, the agent executes action $a_t\in \mathcal{A}$ based on the environment state $s_t\in \mathcal{S}$. Then, the environment yields  an extrinsic reward $r^e_t(s_t,a_t)$ to the agent and makes a transition to a next state $s_{t+1}$ according to  the reward function $R$ and  the transition probability $\mathcal{T}$, respectively. The agent has a policy $\pi$ and aims to find an optimal policy that maximizes the expected return ${\mathbb{E}}[G_0]$, where $G_t=\sum_{k=0}^{\infty}\gamma^k r^e_{t+k}$ is the discounted return at time step $t$.

\textbf{Exploration in RL} ~~ Exploration is a crucial aspect of RL as it allows the agent to gather information about the environment to improve its decision-making ability. Insufficient exploration can result in suboptimal policies.  Simple exploration strategies  include $\epsilon$-greedy \cite{van2016deep}, noise injection \cite{lillicrap2015continuous}, and entropy regularization \cite{schulman2017proximal}. These strategies are still  adopted commonly due to their simplicity and versatility \cite{dabney2020temporally}. However, their limited inductive bias towards transitions under the current policy limits their use to hard exploration tasks in which significant deviations from the learned policy are required.  To address this limitation, several other approaches such as  intrinsic motivation-based exploration \cite{bellemare2016unifying, achiam2017surprise, burda2018exploration} and temporally-extended exploration \cite{osband2016deep, dabney2020temporally, yu2021taac} have been proposed.

\textit{Intrinsic motivation-based exploration} is based on adding an additional bonus, called intrinsic reward, to the extrinsic reward from the environment for better exploration. Intrinsic rewards in existing works are basically  designed based on the `curiosity' of state capturing state visitation frequency so that less-visited states are assigned  higher intrinsic rewards and frequently-visited states are assigned lower intrinsic rewards. One specific approach to designing the intrinsic reward function is the count-based exploration method, which directly exploits the visitation frequency to determine  new states. \citet{bellemare2016unifying} proposed using a density model to approximate the visitation count and utilized it as an exploration bonus. Another approach is the prediction-based exploration method, which measures the curiosity of state based on the error of prediction of the output of  the environment model \cite{achiam2017surprise, stadie2015incentivizing} or a randomly-fixed network \cite{burda2018exploration}. The  rationale behind this approach is as follows.  By using a predictor  that is well-trained  with  sample trajectories, the prediction error on   state-action pairs frequently-observed  in the sample trajectories is small, whereas that on  state-action pairs less-observed in the sample trajectories is large. 
By giving large intrinsic rewards to less-observed state-action pairs, we can encourage the agent to explore less-visited uncertain state-action pairs. 
Random Network Distillation (RND) \cite{burda2018exploration} uses the mean square   error (MSE), $\| \hat{f}(s_t;\theta_{RND}) - f(s_t) \|^2$, of  a neural-network estimator $\hat{f}(s_t;\theta_{RND})$ predicting 
the output of a randomly-initialized fixed  neural network $f(s_t)$   as the intrinsic reward, where the estimator $\hat{f}(s_t;\theta_{RND})$ is trained to predict the output of $f(s_t)$ based on  the collected experiences. The RND-based intrinsic reward 
 is shown to be effective for hard exploration tasks.

\textit{Temporally-extended exploration} refers to the concept of exploration over an extended period of time rather than exploring the state-action space at each time step independently. Several methods have been proposed to leverage this concept in order to  enhance simple exploration strategies. For example, \citet{osband2016deep} proposed a bootstrapping method with multiple heads of deep Q-network to leverage uncertainty estimates for better exploration.  \citet{yu2021taac} proposed a two-stage policy which allows the agent to choose between new actions and the previous actions in order to achieve close-loop temporal abstraction. \citet{dabney2020temporally} introduced a temporally-extended variant of $\epsilon$-greedy named $\epsilon z$-greedy, which first selects a random action and then repeats the chosen action for a duration generated by a zeta distribution, rather than choosing a random action at every time step with probability $\epsilon$.

\textbf{Option Framework} ~~~The option framework provides a method to learn temporally-extended sequences of actions, referred to as options \cite{sutton1999between,bacon2017option}. Option is  a generalization of the concept of action, representing high-level behavior composed of multiple sub-actions. Each option $\omega \in \Omega$ consists of three components: $\pi_{w}$, $\mathcal{I}_{w}$, and $\beta_{w}$, where $\Omega$ is the set of options, $\pi_{w}$ is an intra-option policy (we will simply call intra-policy), $\mathcal{I}_{w} \subseteq \mathcal{S}$ is an initiation set defining the set of states on which option $\omega$ is available, 
and $\beta_{w} : \mathcal{S} \rightarrow [0, 1]$ is a termination function which generates the termination probability. In this paper, we assume $\mathcal{I}_{w}=  \mathcal{S},\forall w \in \Omega$.

To implement algorithms under the option framework, the \textit{call-and-return} option execution model has  been  commonly  adopted \cite{bacon2017option, klissarov2021flexible}. 
In this model,  the agent selects an option $w$ according to an option selection policy $\pi_{\Omega}$ defined over  $\Omega$.  Then, the selected option determines 
 the intra-policy $\pi_{w}$ and the termination function $\beta_{w}$. 
At each time step $t$, based on the chosen option $w$,   the agent selects action $a\sim \pi_{w}(\cdot|s)$ with the determined intra-policy $\pi_{w}$. The action yields the next state $s_{t+1}$ and then the termination function  decides whether  the current option is terminated or not according to the termination probability  $\beta_\omega (s_{t+1})$.  If termination is decided, a new option is selected at the next time step according to the option selection policy $\pi_\Omega$. Otherwise, the current option is continued at the next time step. Then, the action for time step $t+1$ is drawn by the intra-policy determined by the option at time step $t+1$, and the process repeats.

\section{Methodology}

The most effective exploration strategy varies over task and learning phase. To construct a unified exploration strategy that is universal across tasks and learning phases, we consider multiple component  exploration strategies each of which has a certain advantage in exploration, and integrate these component exploration strategies in a single combined exploration strategy. Then, we make the  integrated strategy select the most effective component exploration strategy adaptively over the training phase for a given task from the context of exploration-exploitation trade-off.  The key challenge here is that we do not know which component strategy is most effective at each phase of training, but  need to learn this adaptive selection over time. The proposed unified exploration method achieves this with the option framework, targeting widely-used action-value methods for model-free off-policy RL. Thus, we name the proposed  method LESSON, abbreviating    \textbf{L}earning to integrate the component \textbf{E}xploration \textbf{S}trategie\textbf{S} with an \textbf{O}ptio\textbf{N} framework.  

The key ideas of LESSON are as follows:  

1) We  separate the behavior policy and the target policy, and replace  the behavior policy  with a call-and-return option model mentioned in the previous subsection.  

2) We set the $N$ intra-policies of the call-and-return option model as the greedy policy and  $N-1$ component exploration strategies.  

3) Then, we train train both the option model and the target policy with their respective  objectives based on the trajectories generated by  the option-based  behavior policy.

\begin{figure}[t]
\begin{center}
\begin{tabular}{c}
      \includegraphics[width=0.49\textwidth]{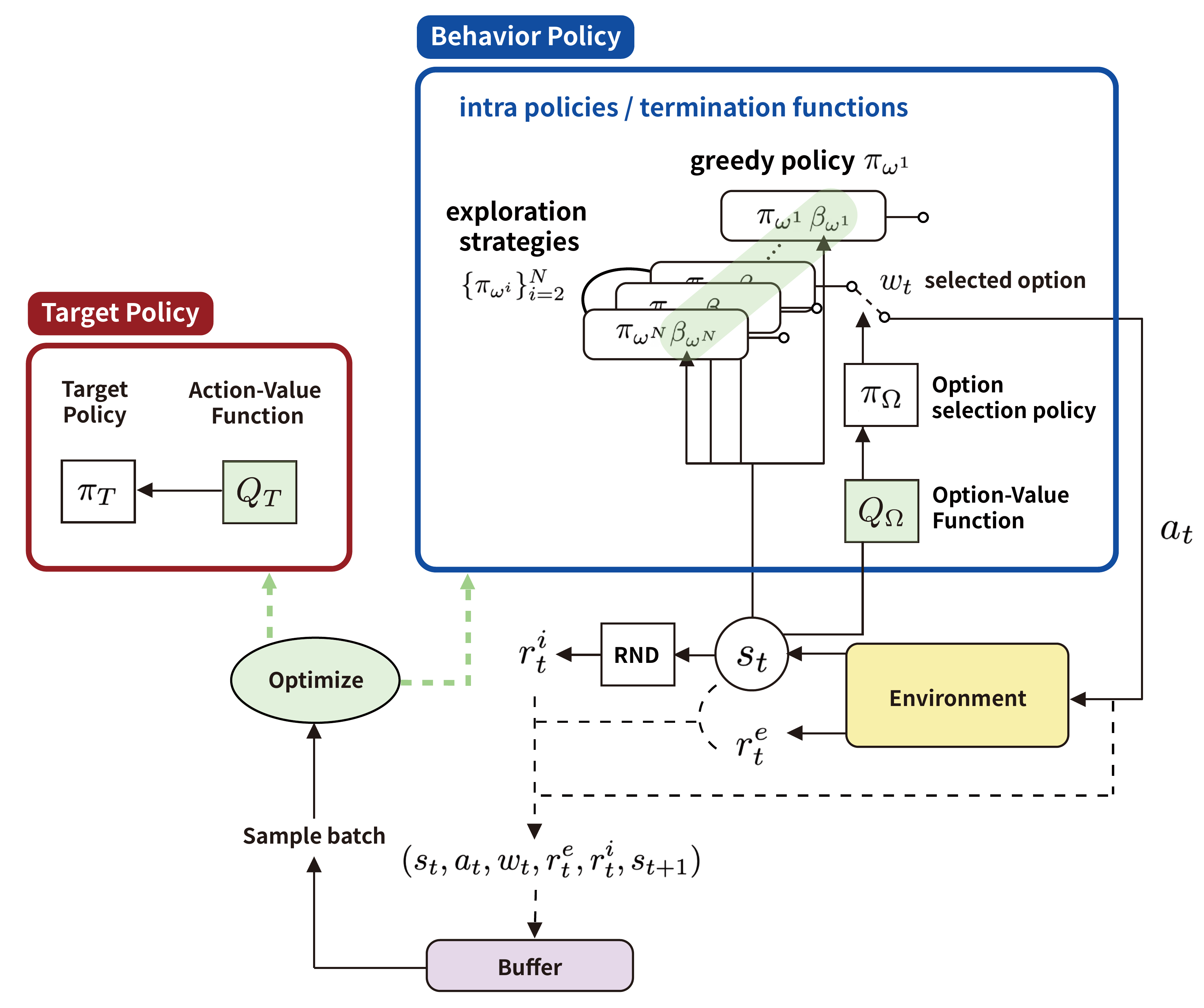}
\end{tabular}
\end{center}
\vspace{-1em}
\caption{Overall diagram of  LESSON: The blue box shows the behavior policy realized by the proposed option  model. The option selection policy $\pi_\Omega$ selects an intra-policy and the corresponding termination function. The target policy denoted by the red box is trained using the samples generated by the behavior policy.}
\label{fig:architecture}
\end{figure}

The overall architecture of LESSON is shown in Fig. \ref{fig:architecture}.   
The inclusion of the greedy policy as one of the intra-policies of the option model is crucial because this inclusion enables  exploration-exploitation trade-off in  sampling. Furthermore, the design of the objective function for the option-based  behavior policy is also important because this objective function implements the exploration-exploitation trade-off in  sampling.  The details follow in the upcoming subsections.

\subsection{Target Policy}

First, we construct the target policy. For the target policy, we consider the widely-used deep Q-network (DQN) \citep{mnih2015human}. The target policy learns  to maximize the expected sum of pure extrinsic rewards, which is the ultimate goal of RL. For this, we define the target action value function as 
\begin{equation}\label{eq:target_q_function}
    Q_T(s_t, a_t) = {\mathbb{E}}\left[\sum_{l=t}^{\infty}\gamma^{l-t}r^{e}_{l} \Big|s_t, a_t\right].
\end{equation}
Then, the target policy is given by the greedy policy 
\begin{equation} \label{eq:targetpolicyfinal}
\pi_T(s) = \mathop{\arg\max}_a Q_T(s,a).
\end{equation}
The target action value function is 
trained to minimize the square of temporal difference (TD) error \citep{mnih2015human}:
\begin{align}\label{eq:loss_dqn}
    \mathcal{L}(\theta) &= {\mathbb{E}}_{(s_l,a_l,r_l,s_{l+1}) \sim \mathcal{D}}\big[ (y_l^{DQN} - Q_T(s_l,a_l;\theta)^2  \big] \nonumber \\
    &\mbox{where} ~~~ y_l^{DQN} = r_l^e + \gamma \max_{a}Q_T(s_{l+1}, a;\theta^{-}),
\end{align}
$\theta^{-}$ is the parameter of the target network of DQN, and
 $\mathcal{D}$ is  the replay buffer storing the samples generated by the behavior policy.

\subsection{Behavior Policy Construction via Option Model}

In off-policy RL, the behavior policy generates samples for learning whereas  actual control is done for the target policy. For the behavior policy, we employ the option framework composed of  the option selection policy $\pi_\Omega$, $N$ intra-policies $\{\pi_{\omega^i}\}_{i=1}^{N}$, and $N$ termination functions $\{\beta_{\omega^i}\}_{i=1}^{N}$. 
Then, we use the call-and-return option execution model \cite{bacon2017option}.  Unlike the option model  proposed by \citet{bacon2017option} which trains all the  three option components from scratch,  we predefine the $N$ intra-policies $\pi_{\omega^1}, \cdots, \pi_{\omega^N}$, and then train the option selection  policy $\pi_\Omega$ and the termination functions $\beta_{\omega^1}, \cdots, \beta_{\omega^N}$ only.

\textbf{Design of intra-policies} ~~ We choose the  $N$-intra-policies as the greedy policy $\pi_{\omega^1}(s)$ and $N-1$ exploration policies $\pi_{\omega^2}, \cdots, \pi_{\omega^N}$, where the greedy policy $\pi_{\omega^1}(s)$ is equivalent to the target policy $\pi_T(s)$ in \eqref{eq:targetpolicyfinal}  trying to maximize the pure extrinsic reward sum.  One key aspect of our design is that we include the greedy policy $\pi_T=\pi_{\omega^1}$ as an intra-policy. This enables  exploration-exploitation trade-off in sampling by allowing  the behavior policy to visit not only new state-action pairs for exploration but also  the state-action pairs  generated by the greedy policy for exploitation.  Hence, by learning optimal selection of one intra-policy out of the $N$ such intra-policies at each time step, 
we can realize  an effective trade-off between exploitation by the greedy  policy and exploration by $N-1$ exploration policies.

Although $N$ can be general, we choose $N=4$ to incorporate three conspicuous existing exploration strategies:  random policy, temporally-extended (TE) random policy and intrinsically-motivated  policy.
 Each intra-policy is explained in  detail below.  Note that we can include other exploration strategies with a larger $N$ if desired.

{\em  1) The greedy intra-policy} selects the greedy action with respect to (w.r.t.) the action-value function $Q_T(s,a)$ in (\ref{eq:target_q_function}). This policy aims to generate trajectories for exploitation by the target policy. 

{\em  2) Random intra-policy} selects a random action from the set of possible actions $\mathcal{A}$ independently at each time step until the option is terminated. 

{\em  3) TE-random intra-policy} selects a random action from the set of possible actions when this option is chosen and then repeats the selected action until the option is terminated. 

{\em  4) Prediction-error maximizing (PEM) intra-policy } is insprired by intrinsically-motivated exploration and RND \cite{burda2018exploration}.  This intra-policy 
selects the action that maximizes the sum of prediction-error  intrinsic rewards. For this, we construct a separate Q-function, which  estimates the expected sum of  prediction-error intrinsic rewards only: 
\begin{equation}\label{eq:q_int}
    Q_{PE}(s_t,a_t)={\mathbb{E}}\left[\sum_{l=t}^{\infty}\gamma^{l-t}r^{PE}_{l} |s_t, a_t \right] ,
\end{equation}
where $r^{PE}_{t} = \mbox{Normalize}(\|  \hat{f}(s_{t};\theta_{RND})- f(s_{t})\|^2)$ is the normalized  prediction-error intrinsic reward. (Refer to Appendix \ref{appx-rnd-details} for the detail of normalization.) We train $Q_{PE}(s,a)$ by minimizing the one-step TD error with the intrinsic reward and train the prediction network $\hat{f}(s_{l+1};\theta_{RND})$ to follow a  randomly initialized fixed network $f(s_{l+1})$ as in \cite{burda2018exploration}. Then, our PEM intra-policy is given by  $\pi_{\omega^4} = \mbox{argmax}_a Q_{PE}(s,a)$, i.e., it aims to maximize the prediction-error sum only. 
Our PEM intra-policy focuses purely on exploration with ignoring the extrinsic reward, which is a different point from the original RND policy \cite{burda2018exploration}.

Intra-policies 2) and  3) choose  random actions for exploration, where intra-policy 3) is a temporally-extended version 
of intra-policy 2). On the other hand,  intra-policy 4) chooses actions away from those frequently done before based on previous sample trajectories. Thus,   intra-policies 2) and 3) are for {\em sample history-unware random exploration}, whereas intra-policy 4) is for {\em sample history-aware exploration}. Here, we selected PEM intra-policy for sample history-aware exploration, but other sample history-aware exploration such as count-based exploration can also be considered. (See Appendix \ref{sec:appx-montezuma}.)   
We want to mix these two distinct approaches.  Note  that intra-policies 2), 3) and 4) all focus only on pure exploration. When one of these three intra-policies is combined with the greedy policy $\pi_{\omega^1}$, the combination can produce several conventional exploration methods. For example, when we combine the greedy policy $\pi_{\omega^1}$ and the random policy $\pi_{\omega^2}$ with fixed  probabilities $1-\epsilon$ and $\epsilon$, the combination is equivalent to  $\epsilon$-greedy. When we combine the greedy policy $\pi_{\omega^1}$ and the TE-random policy $\pi_{\omega^3}$  
with fixed  probabilities $1-\epsilon$ and $\epsilon$, the combination reduces to $\epsilon$z-greedy for which the TE duration is determined by the option termination period rather than a zeta distribution. Furthermore, the combination of $\pi_{\omega^1}$ and $\pi_{\omega^4}$ yields a similar policy to the RND policy. 
Our goal is not to use such a fixed combination but to learn the most effective combination over time for a given task through the option selection policy $\pi_\Omega$ and the termination functions $\{\beta_\omega\}$.

\subsection{Learning the Option Model}

With the predefined intra-policies, we need to learn the option selection policy $\pi_\Omega$ and the termination functions $\{\beta_\omega\}$ for the behavior policy.  For this, we propose the following objective function for the  behavior policy implemented by the call-and-return option execution model:
\begin{align}\label{eq:obj_optioncritic}
    J(\pi_{\Omega}, \{\beta_{w}\}) = {\mathbb{E}} \left[\sum_{t=0}^{\infty}\gamma^{t}(r^{e}_{t} + ~\alpha r^{i}_{t})\right],
\end{align}
where $r_t^e$ is the extrinsic reward for exploitation and $r_t^i$ is the intrinsic reward for exploration. The reason for this design of the objective function for $\pi_\Omega$ and $\{\beta_\omega\}$ is that the behavior policy should not only sample for exploration but also for exploitation for  a trade-off between these two, where 
 $\alpha$ is the coefficient controlling the trade-off. 
 
For the intrinsic reward for exploration $r_t^i$ in (\ref{eq:obj_optioncritic}), we exploit the existing intrinsic reward of the prediction error proposed by \citet{burda2018exploration} again. That is, we set 
\begin{equation} \label{eq:intrinsic_reward}
r^{i}_{t} := r_{t}^{PE}=\mbox{Normalize}(\|  \hat{f}(s_{t})- f(s_{t})  \|^2),
\end{equation}
where $f(s)$ and $\hat{f}(s)$ are already defined in the part of the PEM intra-policy. Then, the RHS of  (\ref{eq:obj_optioncritic}) is the same as the objective of RND \cite{burda2018exploration}. However, there exists a key difference between our use of (\ref{eq:obj_optioncritic}) and Burda et al.'s   use of (\ref{eq:obj_optioncritic}).
We use (\ref{eq:obj_optioncritic}) for learning the behavior policy while having a separate greedy target policy. In constrast,  \citet{burda2018exploration} use (\ref{eq:obj_optioncritic}) for learning the target policy itself.
Note that (\ref{eq:obj_optioncritic}) without the extrinsic reward $r_t^e$ reduces to the objective  (\ref{eq:q_int}) of the PEM intra-policy. However, the PEM exploration intra-policy is not necessarily selected as the behavior option over the random policies $\pi_{\omega^2}$ and $\pi_{\omega^3}$ due to the extrinsic reward term. This is because the corresponding extrinsic reward resulting from the PEM intra-policy can be smaller than those produced by the random exploration policies 
$\pi_{\omega^2}$ and $\pi_{\omega^3}$, although the PEM exploration intra-policy yields the largest intrinsic reward.  Thus, in our formulation the most effective  intra-policy is selected from the viewpoint of exploration-exploitation trade-off. 

\textbf{Learning Option Selection Policy} ~~To implement the aforementioned call-and-return option execution model, we adopt an option-critic model \cite{bacon2017option}.
 The option-critic model trains the option execution model based on the option-value function. In the case of maximizing  (\ref{eq:obj_optioncritic}), we define  the  option value function as
\begin{align}\label{eq:q_option}
    Q_{\Omega}(s_t, \omega_t) = {\mathbb{E}}\left[\sum_{l=t}^{\infty}\gamma^{l-t}(r^{e}_{l} + \alpha r^{i}_{t}) \big|s_t, \omega_t\right],
\end{align}
where $s_t$ and $\omega_t$ are the state and  option at time step $t$, respectively, and the expectation trajectory follows the described call-and-return option execution model. 
Then, the option selection policy  $\pi_{\Omega}$ is given as the greedy policy w.r.t. $Q_\Omega$, i.e.,
\begin{equation}
\omega_t=\pi_\Omega(s_t)=\mathop{\arg\max}_{\omega'} Q_\Omega(s_t,\omega').
\end{equation}
We parameterize the option-value function $Q_\Omega$ with parameter $\theta_{\Omega}$ and then train it to minimize the one-step TD error loss function: 
\begin{align}\label{eq:loss_policyoveroption}
    \mathcal{L}(\theta_{\Omega})=& {\mathbb{E}}_{(s_t, w_t, r_t^e+\alpha r_t^i,s_{t+1})\sim \mathcal{D}}\big[ (y_t - Q_{\Omega}(s_t,\omega_t;\theta_{\Omega}))^2 \big], 
\end{align}
where \cite{bacon2017option}
\begin{align}\label{eq:loss_policyoveroption2}
    y_t&=r_t^e + \alpha r_t^i + \gamma \Bigg(\Big(1-\beta_{\omega_t}(s_{t+1})\Big)Q_{\Omega}(s_{t+1}, \omega_t;\theta^{-}_{\Omega}) \nonumber \\
    & ~~~~~~~~~~~~~~~~~~ +\beta_{\omega_t}(s_{t+1})\max_{w'}Q_{\Omega}(s_{t+1}, w';\theta^{-}_{\Omega})\Bigg).
\end{align}
Here, $\mathcal{D}$ is the replay buffer, and $\theta^{-}_{\Omega}$ is the parameter of the target option-value network. For the first term in the big  parenthesis of the right-hand side (RHS) of   (\ref{eq:loss_policyoveroption2}), $Q_{\Omega}(s_{t+1}, \omega_t)$ is used because this corresponds to the case that the current option $\omega_t$ is not terminated at $t+1$ with probability $1-\beta_{\omega_t}(s_{t+1})$. The second term in the parenthesis corresponds to the case that the current option is terminated with probability  $\beta_{\omega_t}(s_{t+1})$ at $t+1$ and a new option is selected according to the greedy policy w.r.t. $Q_{\Omega}(s_{t+1}, \cdot|\theta_\Omega^-)$.

\begin{algorithm}[t]
\SetAlgoLined
Initialize target action-value function $Q_{T, \theta}$, option-value function $Q_{\Omega, \theta_{\Omega}}$, termination functions  $\{\beta_\omega\}$, intrinsic reward coefficient $\alpha$, replay buffer $\mathcal{D}$ 

Choose $\omega_0$ according to  option selection policy $\pi_{\Omega}(s_0)$
 
 \For{each iteration}{
    \For{each environment time step $t$}{
      
      Observe $s_t$ and       choose  $a_t$ according to intra-policy $\pi_{\omega_t}(a_t|s_t)$
      
      Take action $a_t$ and  receive  $r^e_t$
        and $s_{t+1}$
        
      Calculate intrinsic reward $r^i_t$ via \eqref{eq:intrinsic_reward} 
      
      Store $(s_t, a_t, w_t, r_t^e, r_t^i, s_{t+1})$ in replay buffer ${\mathcal D}$ 
      
      \uIf{$\beta_{\omega_t}$ decides termination} {
        Choose new $\omega_{t+1}$ according to  $\pi_{\Omega}(s_{t+1})$
      }
      \Else{
        $\omega_{t+1}\leftarrow \omega_t$
      }
   }
   \For{each update time step $t$}{
    Sample $e_t = (s_t, a_t, w_t, r_t^e,  r_t^i, s_{t+1})$ from ${\mathcal D} $

    {Update target action-value function} $Q_T$ by applying SGD  to the loss ${\mathcal L}(\theta)$ in \eqref{eq:loss_dqn} 
    
    {Update option-value function} $Q_\Omega$  by applying SGD to the loss $\mathcal{L}(\theta_{\Omega})$ in \eqref{eq:loss_policyoveroption}

    {Update termination functions} $\{\beta_\omega\}$ by using the gradient \eqref{eq:grad_termination}
   }
 }
\caption{LESSON}
\label{alg:LESSON}
\end{algorithm}

\textbf{Learning Termination Functions} ~~ The termination function generates the termination probability of the associated option. To update the termination functions, we use  the gradient of the option-value function $Q_\Omega$ w.r.t. to the parameter $\theta_{\beta_\omega}$ of the termination function for each option, which is given by \citep{bacon2017option}
\begin{align}
 \frac{\partial Q_{\Omega}}{\partial \theta_{\beta_\omega}}&=   -{\mathbb{E}} \big[\nabla_{\theta_{\beta_\omega}}\beta_{\omega}(s_{t+1};\theta_{\beta_\omega}) \times \nonumber \\
    &~~~~~~~~~~~~~(Q_{\Omega}(s_{t+1}, \omega_{t})-\max_{\omega'} Q_{\Omega}(s_{t+1}, \omega'))\big]. \label{eq:grad_termination}
\end{align}

Note that the gradient form (\ref{eq:grad_termination}) is similar to that of the conventional policy gradient \citep{sutton2018reinforcement}. 
Due to the form of the gradient in (\ref{eq:grad_termination}), if an option is not optimal at time step $t+1$, then the  advantage $Q_\Omega (s_{t+1},\omega_t)-\max_{\omega'} Q_\Omega (s_{t+1},\omega')$ becomes negative, the termination probability of that option is  trained to increase.  Consequently, options with low option values are more likely to be terminated quickly, while those with high option values are more likely to be retained. 

\begin{figure*}[t!]
\begin{center}
\includegraphics[width=0.75\textwidth]{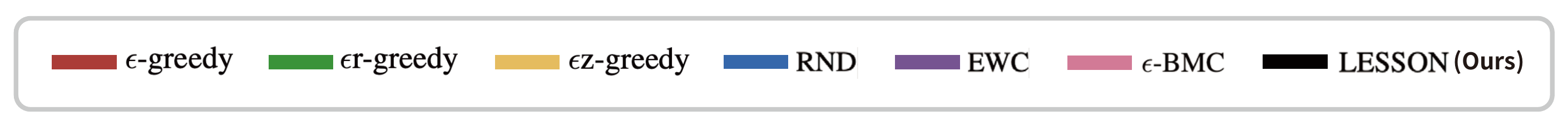}
\begin{tabular}{cccc}
      \includegraphics[width=0.23\textwidth]{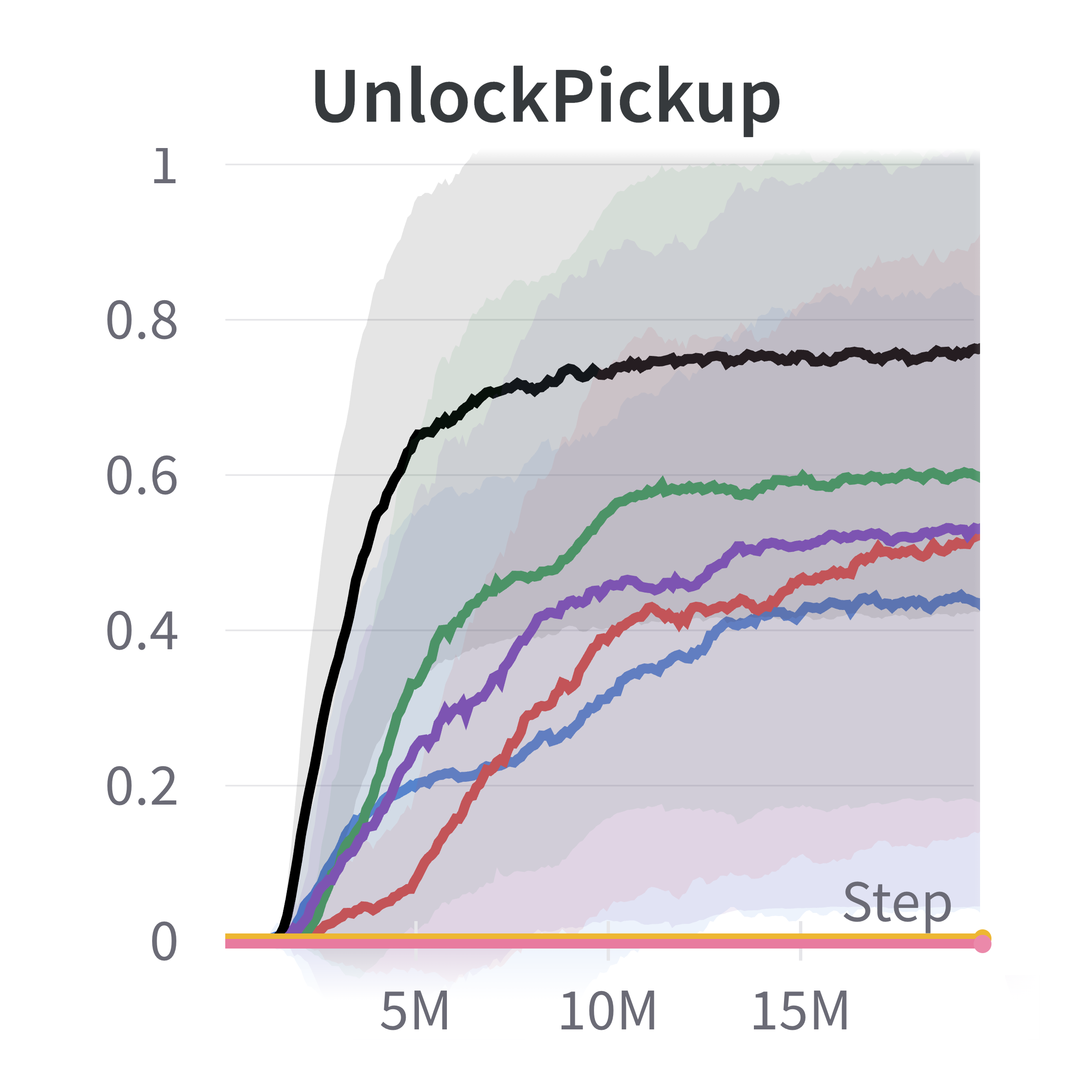} &
      \includegraphics[width=0.23\textwidth]{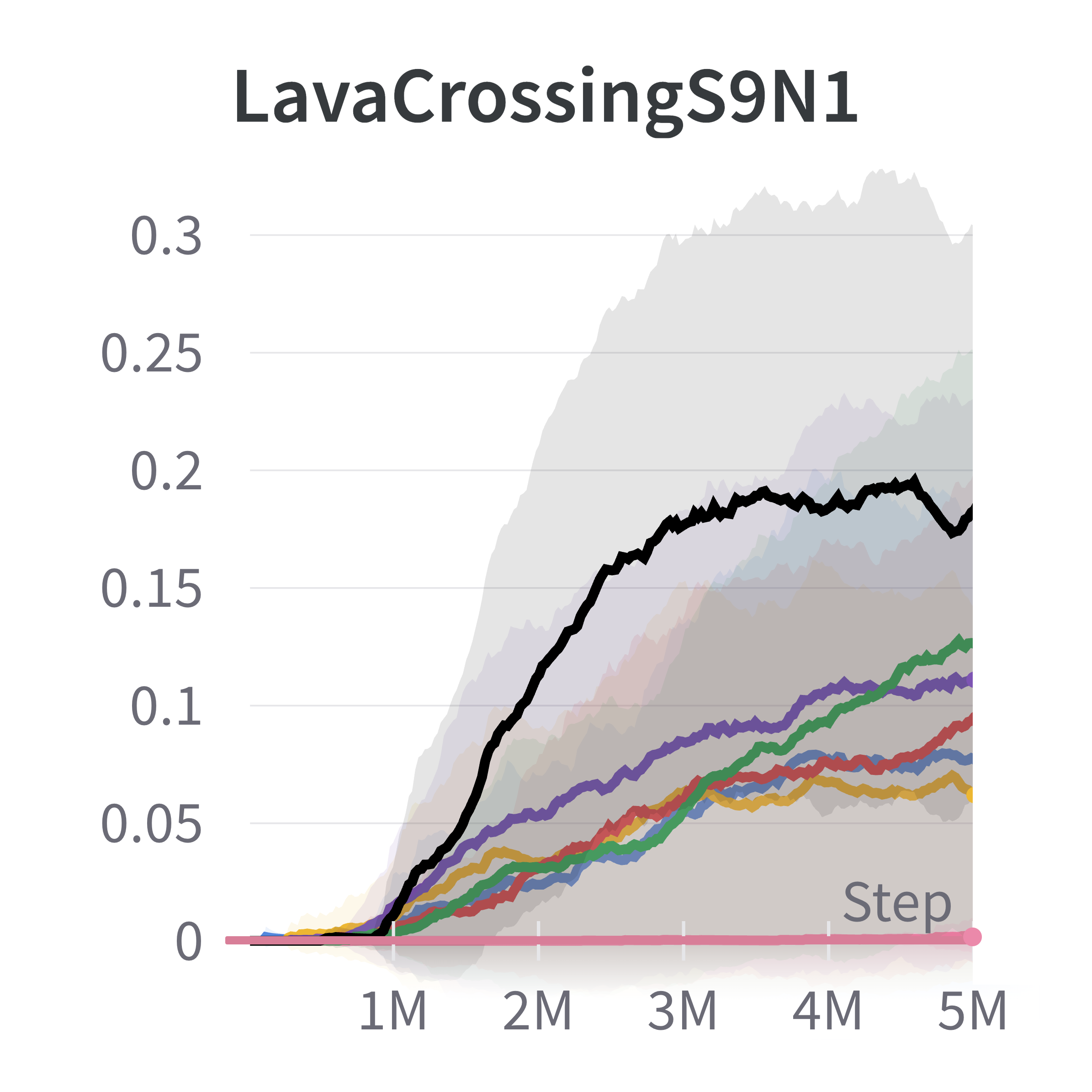} & 
      \includegraphics[width=0.23\textwidth]{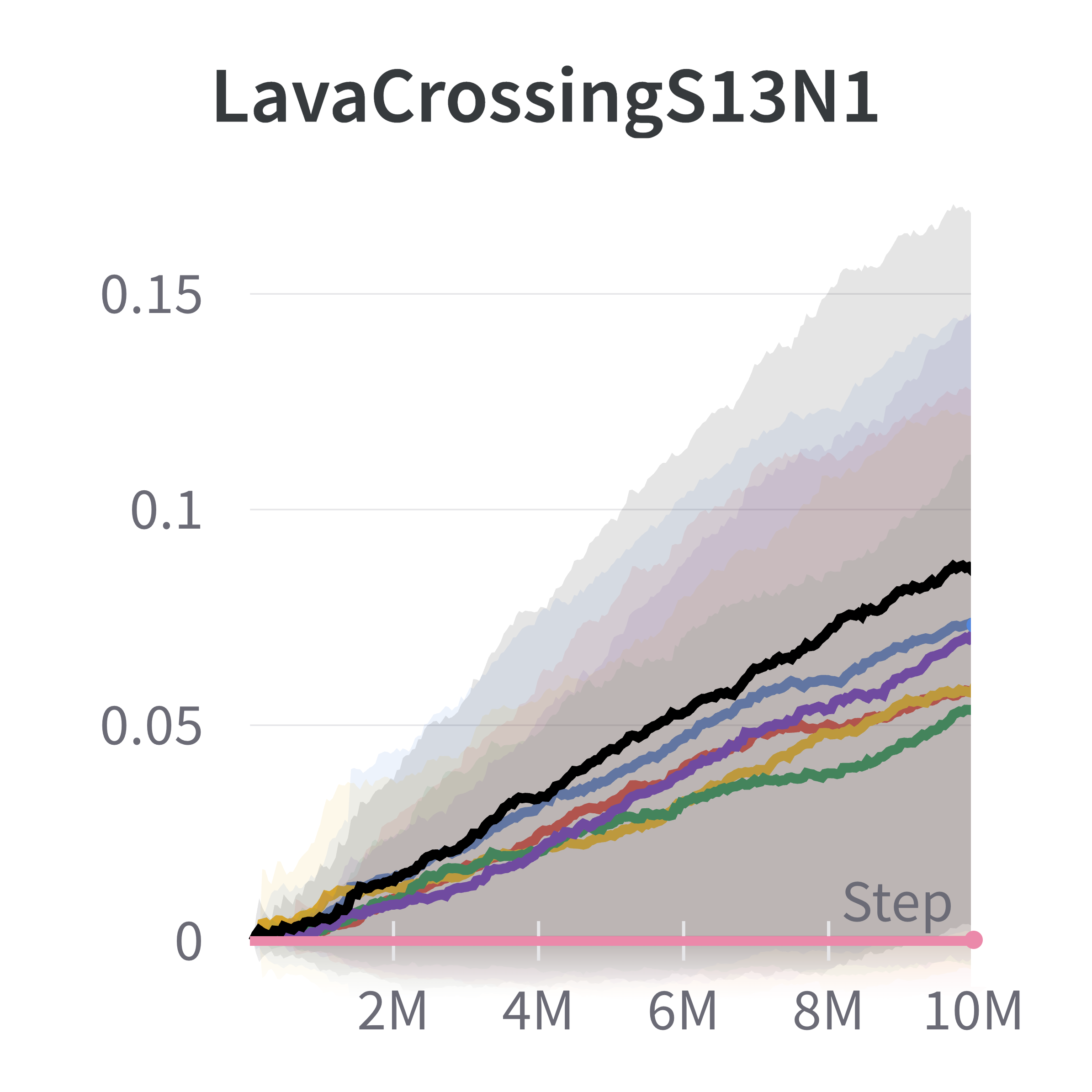} & 
      \includegraphics[width=0.23\textwidth]{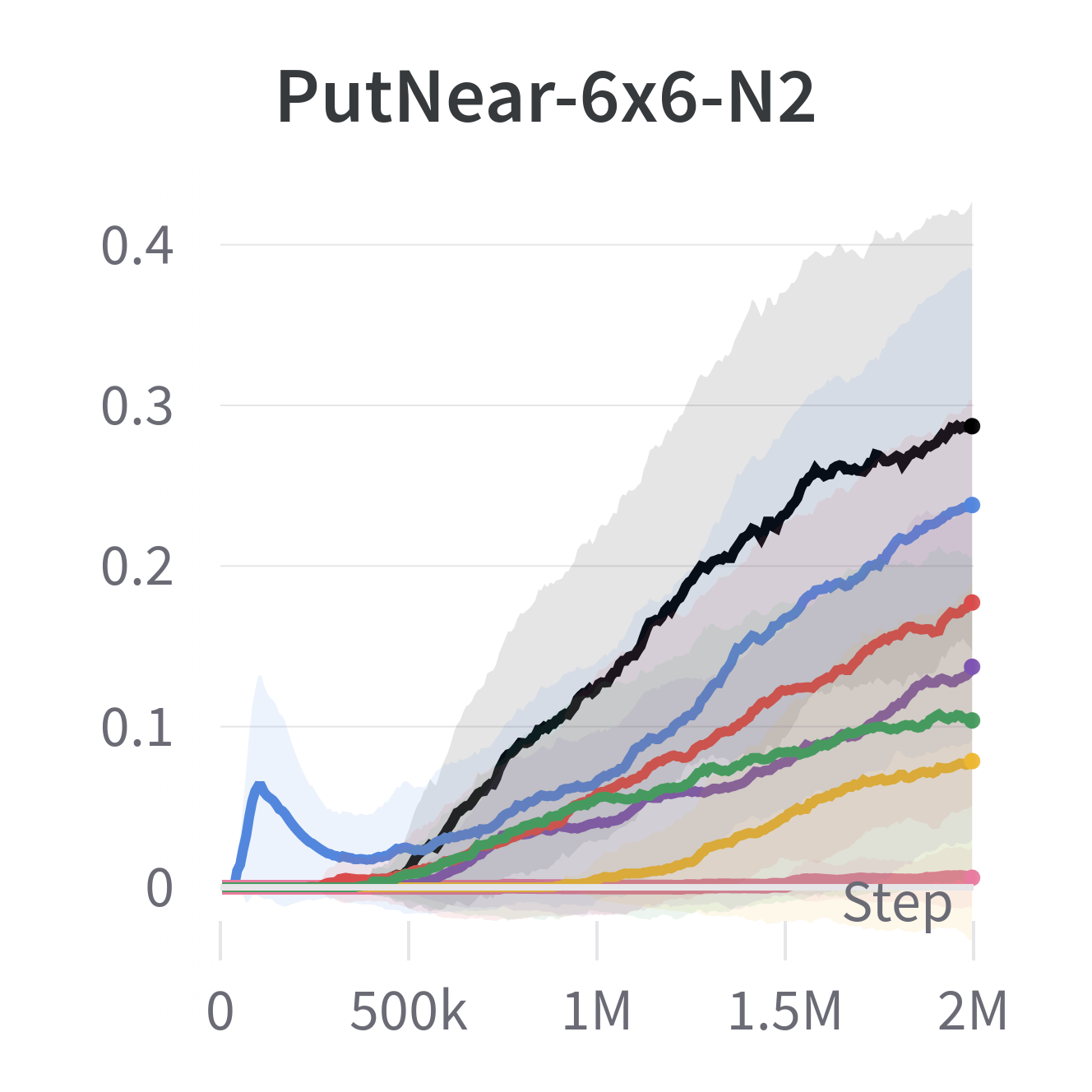} \vspace{-1ex} \\
      (a) & (b) & (c) & (d) \\
      \includegraphics[width=0.23\textwidth]{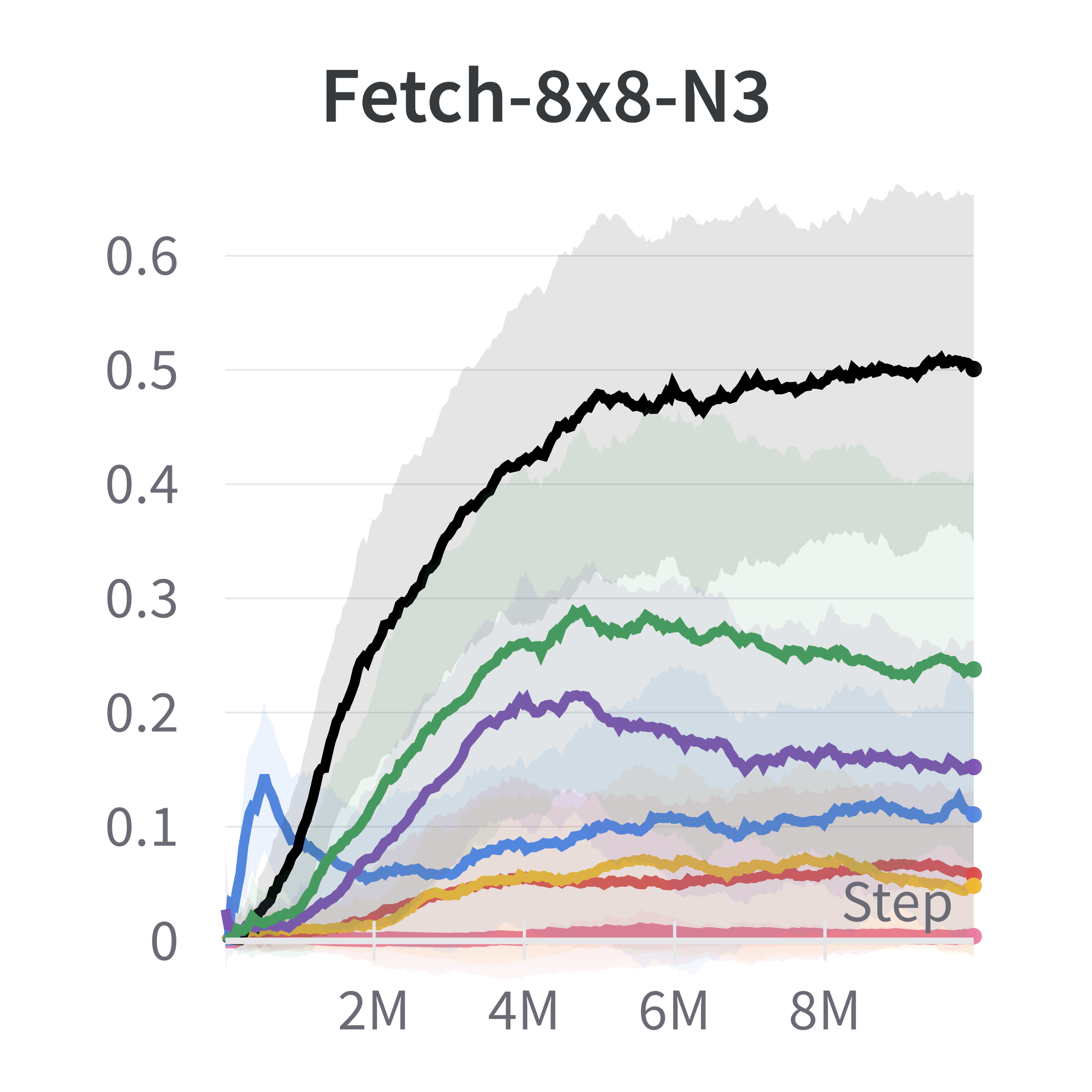} &
      \includegraphics[width=0.23\textwidth]{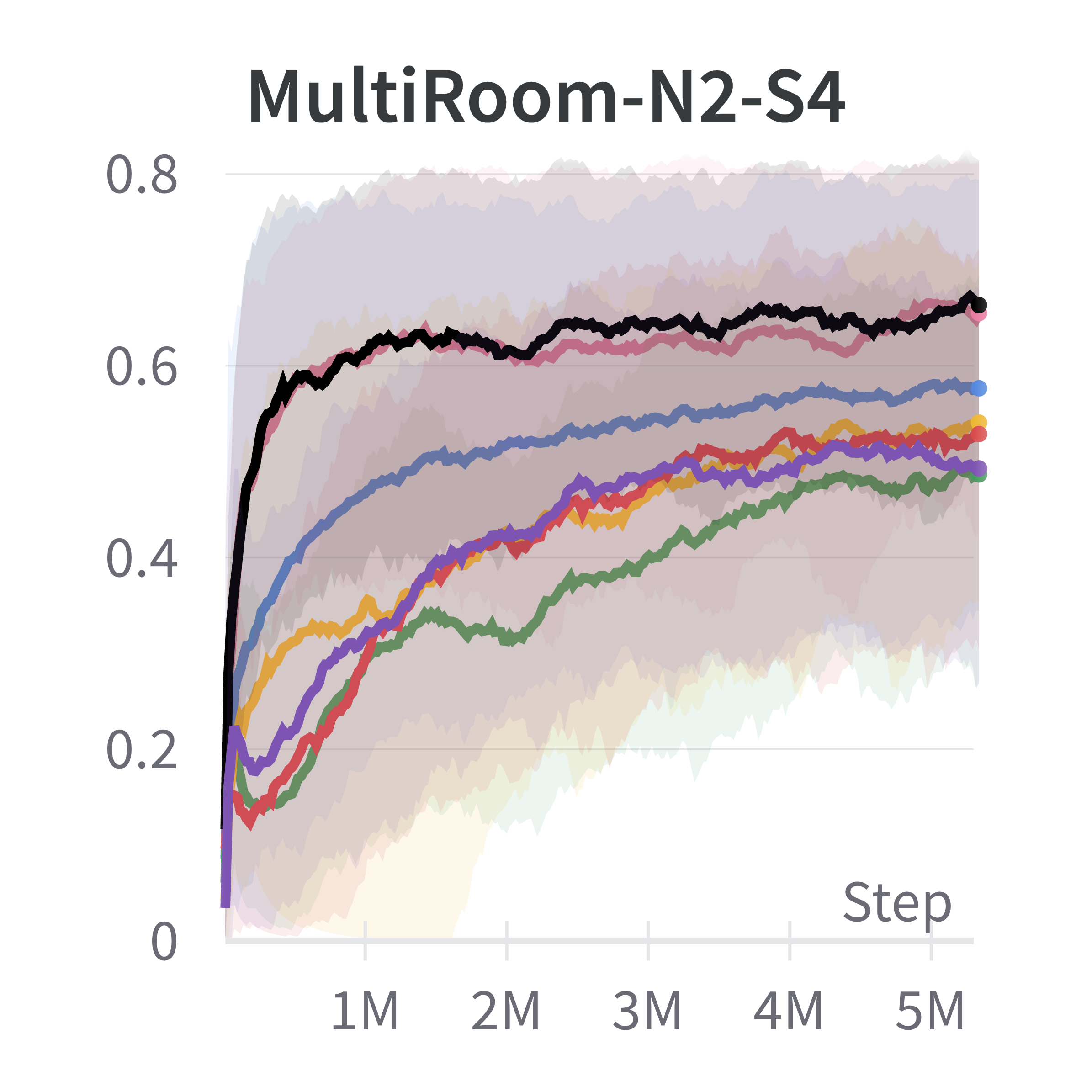} &
      \includegraphics[width=0.23\textwidth]{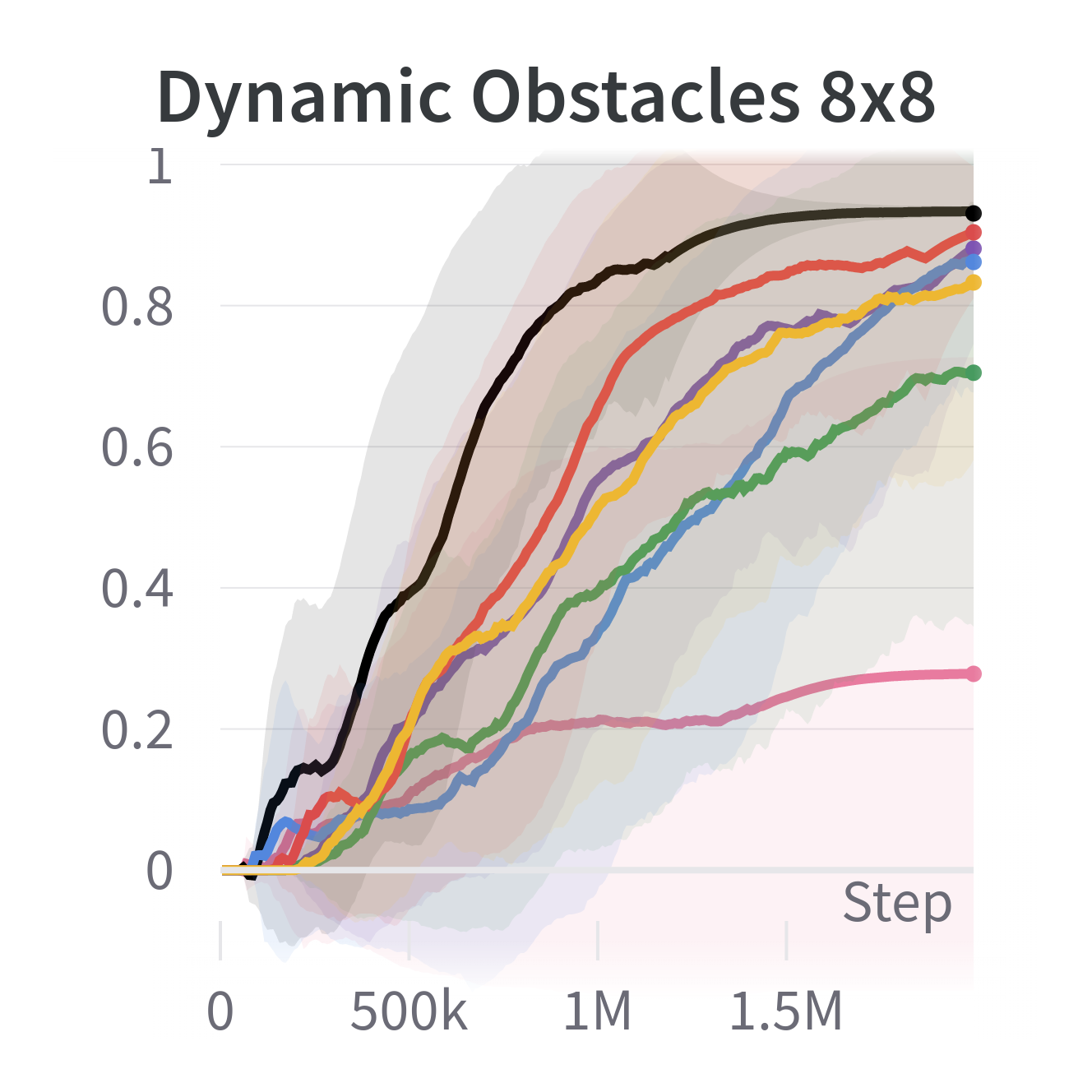} &
      \includegraphics[width=0.23\textwidth]{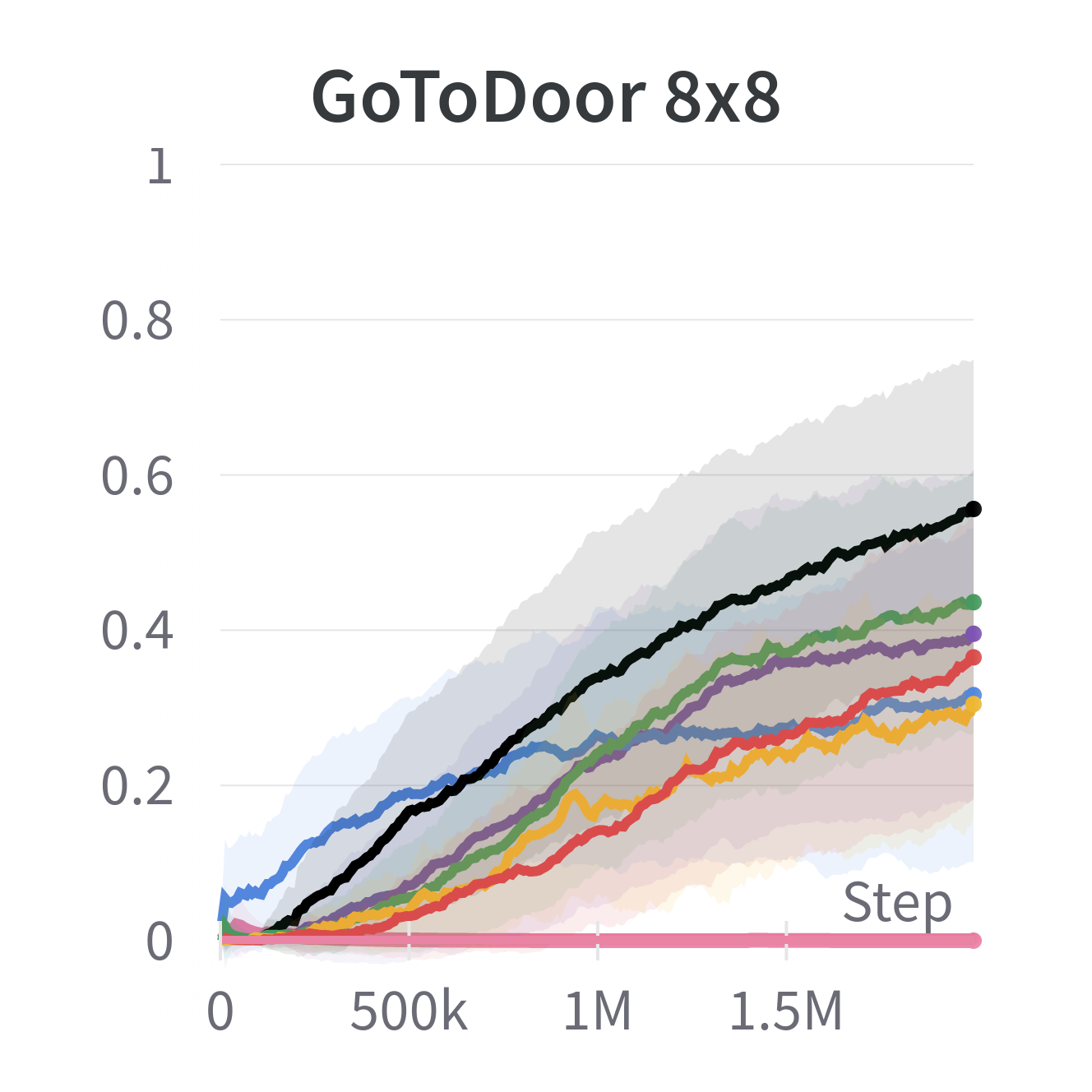}
      \vspace{-1ex}\\
      (e) & (f) & (g) & (h)
\end{tabular}
\end{center}
\caption{Performance comparison on the MiniGrid tasks. More results are provided in Appendix \ref{sec:appx-experimental-results}.}
\label{fig:OverallResults}
\end{figure*}

\begin{figure*}[t!]
\begin{center}
\includegraphics[width=0.75\textwidth]{figure/label/label.pdf}
\begin{tabular}{cccc}
      \includegraphics[width=0.23\textwidth]{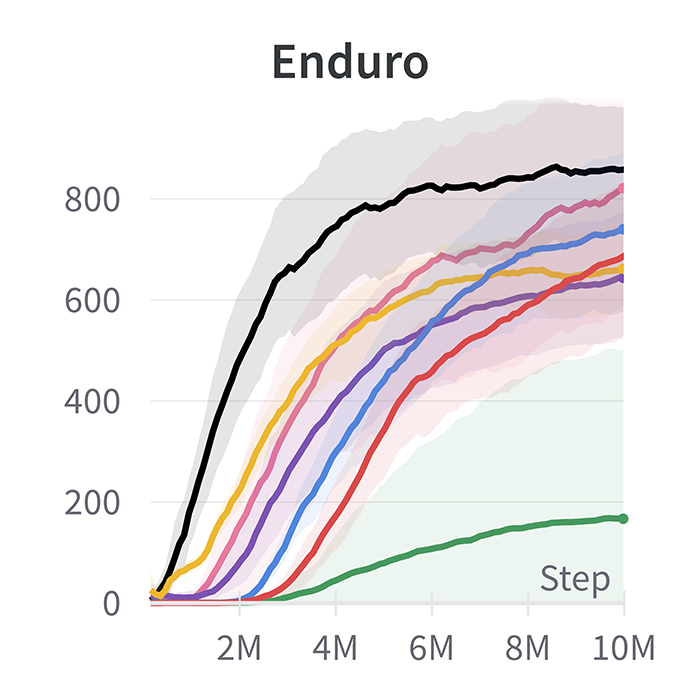} &
      \includegraphics[width=0.23\textwidth]{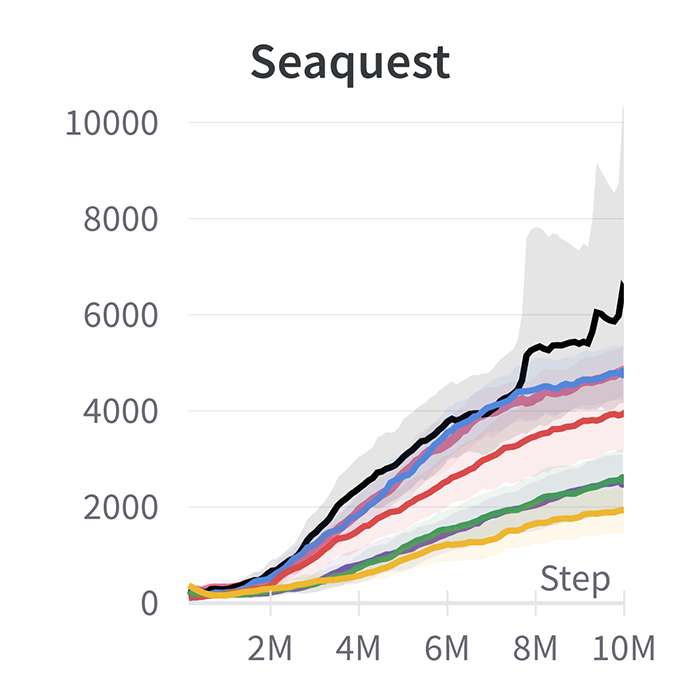} &
      \includegraphics[width=0.23\textwidth]{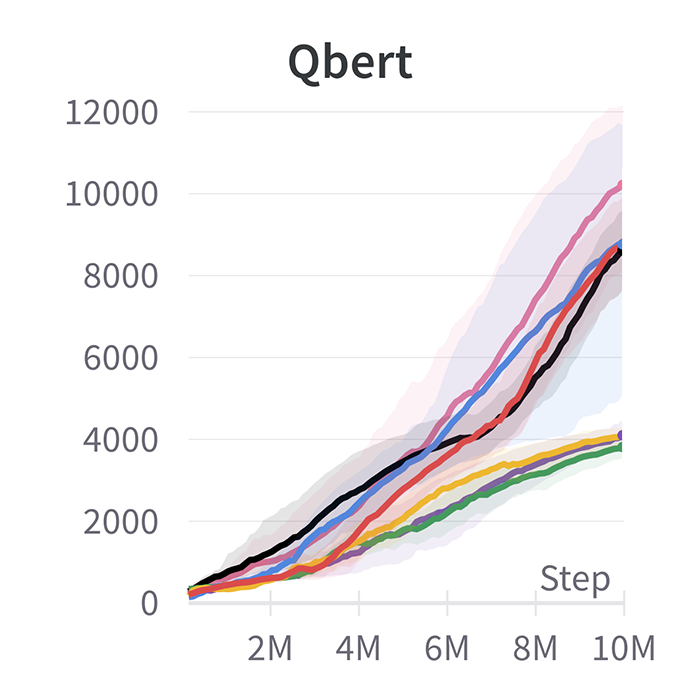} &
      \includegraphics[width=0.23\textwidth]{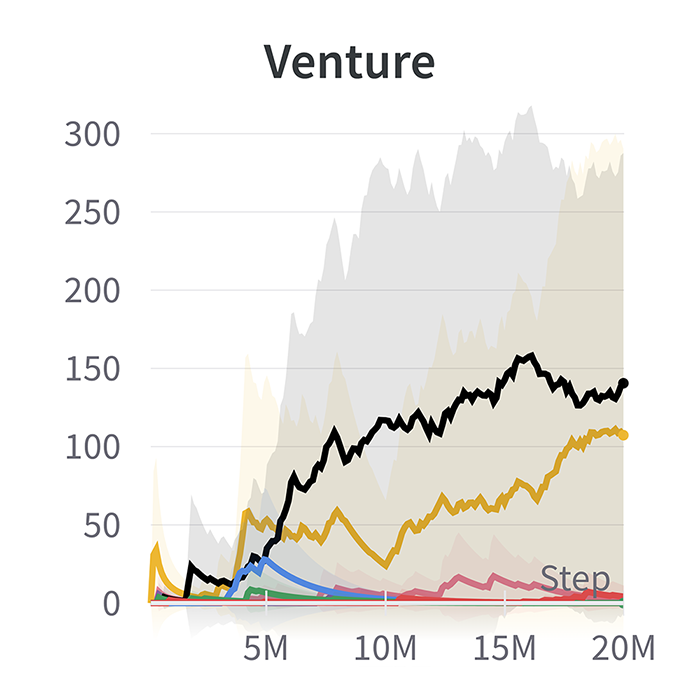}
      \vspace{-1ex}\\
      (a) & (b) & (c) & (d)
\end{tabular}
\end{center}
\caption{Performance comparison on the Atari 2600 tasks}
\label{fig:AtariResults}
\end{figure*}

Note that we fixed $\alpha$ controlling the exploration-exploitation trade-off in the objective (\ref{eq:obj_optioncritic}) for sampling. However, the fixed $\alpha$ does not mean the trade-off between exploration and exploitation is fixed over time for a given task because the termination functions for the greedy and exploration policies are learned and updated over time. Thus, a proper time-varying exploration-exploitation trade-off can be learned over time for given $\alpha$, as we will see in Section \ref{sec:experiments}. The final algorithm of LESSON is summarized in Algorithm  \ref{alg:LESSON}. The  implementation is based on  the ideas explained here, but some minor implementation detail is added. The software code of LESSON is available at \url{https://github.com/beanie00/LESSON}.

\section{Experiments}\label{sec:experiments}

\begin{figure*}[t!]
    \centering
    \begin{tabular}{cccc}
      \raisebox{0.3\height}{\includegraphics[width=0.12\textwidth]{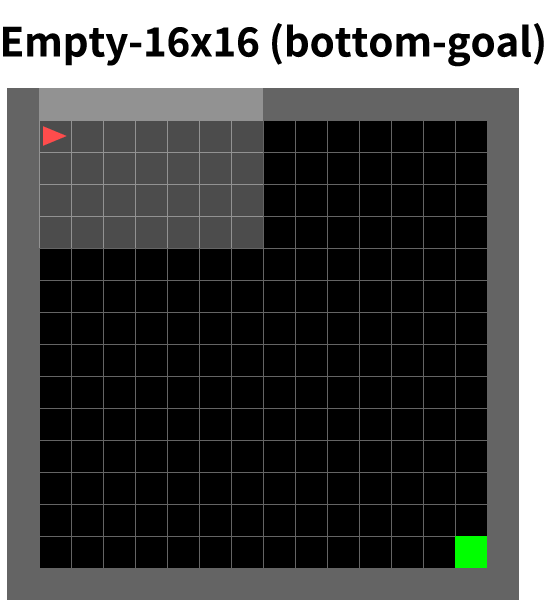}} \hspace{-2ex}&
      \raisebox{.0\height}{\includegraphics[width=0.17\textwidth]{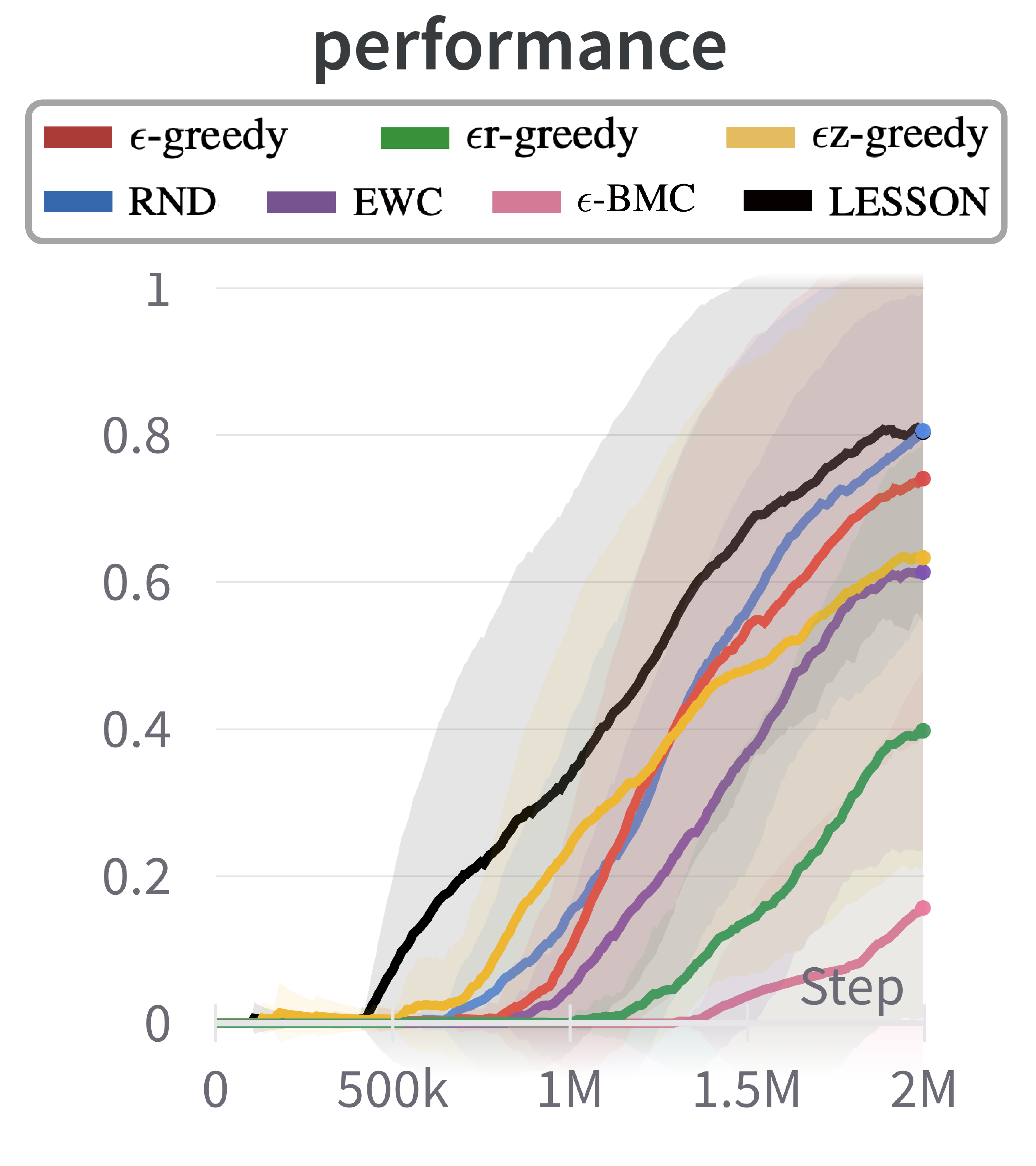}} & 
      \hspace{-2ex}
      \raisebox{.0\height}{\includegraphics[width=0.17\textwidth]{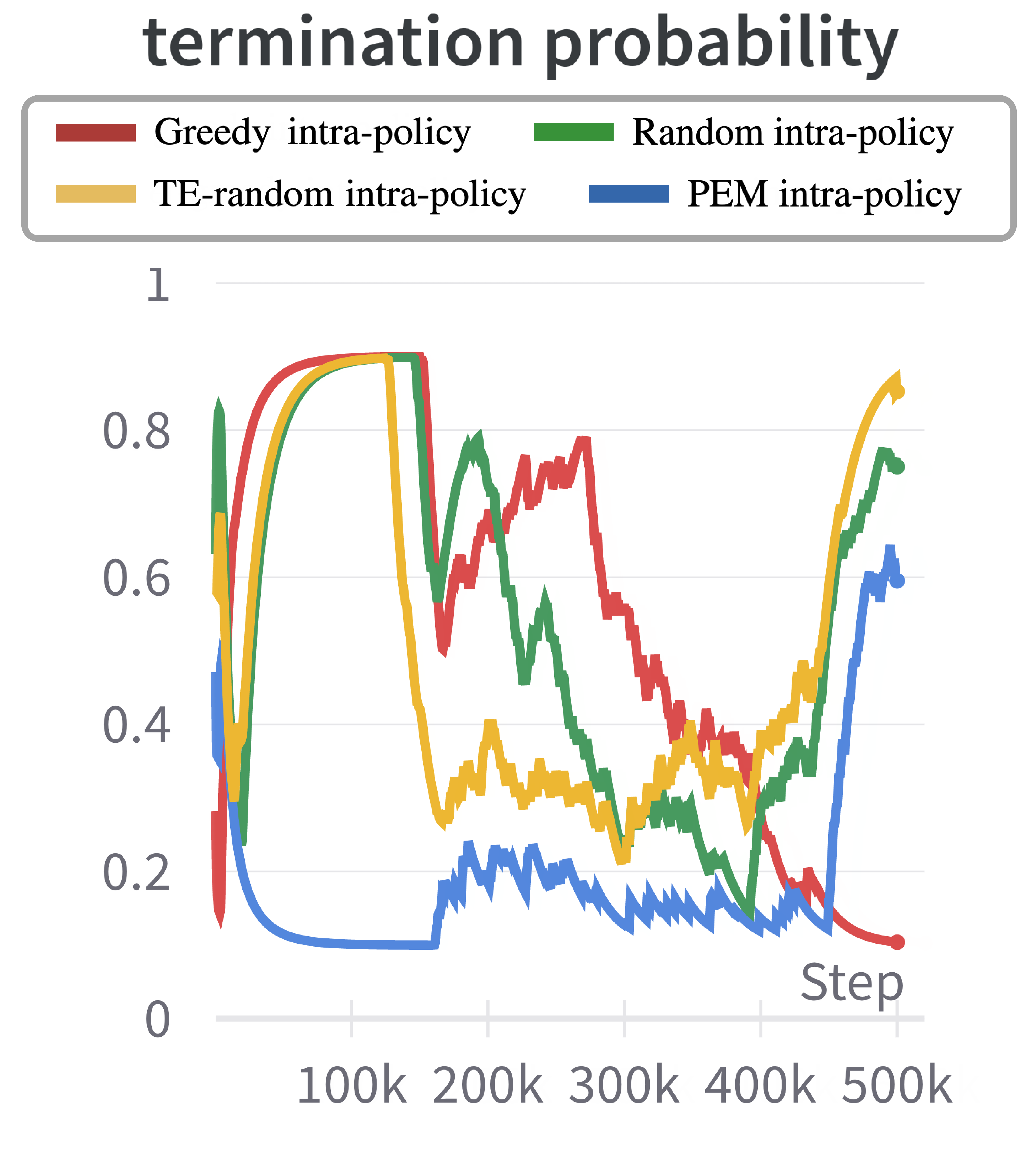}} \hspace{-2ex}& 
      \includegraphics[width=0.47\textwidth]{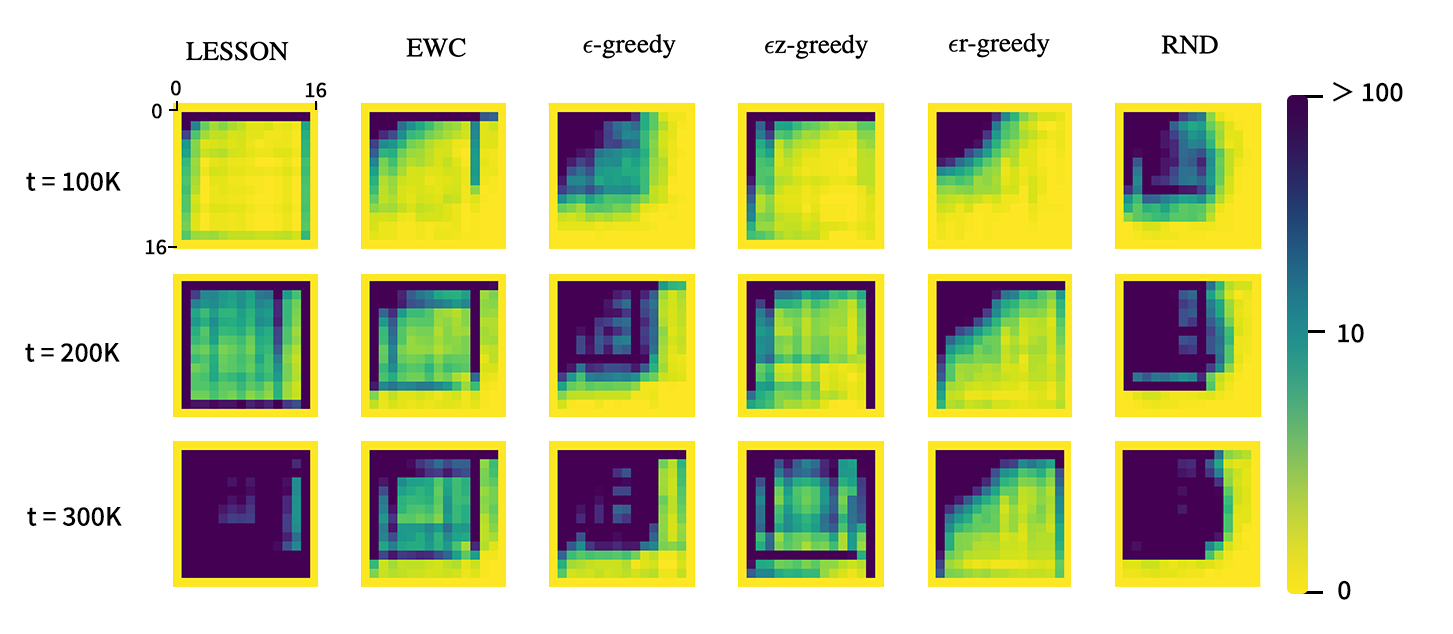} \\
      (a) & (b) & (c) & (d)
    \end{tabular}
    \vspace{-1.5em}
\caption{Comparison of LESSON with the baselines in the Empty-16x16 environment with the goal at the right lower corner: (a) the view of environment, (b) performance comparison, (c) the termination probabilities $\beta_\omega$ over time for LESSON, and (d) state visitation frequency. (Fig.4(a) was obtained by rendering the MiniGrid Empty-16x16  environment while training \cite{gym_MiniGrid}.)}
\label{fig:visitation}
\end{figure*}

To evaluate the performance of  LESSON, we compared it with  major  baselines. All algorithms are based on DQN \cite{mnih2015human} but employ different exploration strategies. The baselines are  1) $\epsilon$-greedy: vanilla  DQN,  2) $\epsilon$z-greedy: if exploration is decided at a time step with probability $\epsilon$, it generates the duration of action-repeat using a zeta distribution and then repeats the same action for the determined duration,  3) $\epsilon$r-greedy: if exploration is decided at a time step with probability $\epsilon$, it generates a duration using a zeta distribution and then selects  independent random actions  during the duration, 
4) RND-based DQN (RND) \cite{burda2018exploration} adds  RND-based intrinsic reward to extrinsic reward, and uses $\epsilon$-greedy DQN, 5) Equal weight combining (EWC) selects one of the previous four exploration strategies (1) - (4) randomly  with equal probabilities, and 6) $\epsilon$-BMC \cite{gimelfarb2020epsilon} is 
 $\epsilon$-greedy DQN, where $\epsilon$ is learned to find a good exploration-exploitation trade-off over time. 

We evaluated the algorithms on fourteen MiniGrid environments \cite{gym_MiniGrid} and four Atari 2600 environments \cite{bellemare2012ale}. The
detailed setting of the considered environments is provided in Appendix \ref{sec:appx-environments}. 

\subsection{Performance Comparison} \label{sec:performance}

Figs. \ref{fig:OverallResults} and \ref{fig:AtariResults} show the performance of  LESSON and the baselines on the MiniGrid environments averaged over 10 random seeds and on the Atari 2600 environments averaged over 3 random seeds, respectively. Detailed hyperparameter setting is provided in Appendix \ref{sec:appx-implementation-details}.

It is seen that the best-performing exploration strategy among the baselines varies over  tasks.  In the MultiRoom environment, where the agent should visit other rooms through narrow paths  to reach a goal, RND performs better than other exploration strategies since RND-based intrinsic reward encourages the agent to visit other rooms once it has already visited some rooms. In the case of LavaCrossingS9N1, which involves reaching a goal while avoiding randomly generated obstacles, $\epsilon$r-greedy performs better than RND. However, in the LavaCrossingS13N1, which increases the map size compared with LavaCrossingS9N1, it is seen that RND is more effective than $\epsilon$r-greedy.  In the Venture environment, which aims to find a treasure while fighting with monsters in several rooms, $\epsilon$z-greedy is the best-performing exploration strategy.  These results highlight that the best-performing exploration strategy is affected by the nature of the task including the size of the environment. 
Note that the performance of $\epsilon$-BMC surpasses that of the e-greedy algorithm in Atari environments. However, it does not show comparable performance in most MiniGrid environments. This may be attributed to the sparsity of rewards in MiniGrid tasks, which make it challenging for the $\epsilon$-BMC agent to acquire information about the environment required to learn $\epsilon$.

While the best-performing exploration strategy varies over environments, LESSON  consistently outperforms the baselines in the considered environments since  it integrates exploration strategies to its advantage. In contrast to LESSON, the simple combining of exploration strategies, EWC, tends to perform worse than the best-performing exploration strategy since ineffective exploration strategies are equally used as well as other strategies. Indeed, LESSON provides non-trivial  adaptive integration of  exploration strategies. 

More results on other MiniGrid tasks are provided in Appendix \ref{sec:appx-experimental-results} and the result on Atari's Montezuma's Revenge is separately provided in Appendix \ref{sec:appx-montezuma}.

\begin{figure*}[hbt!]
  \begin{center}
  \includegraphics[width=0.7\textwidth]{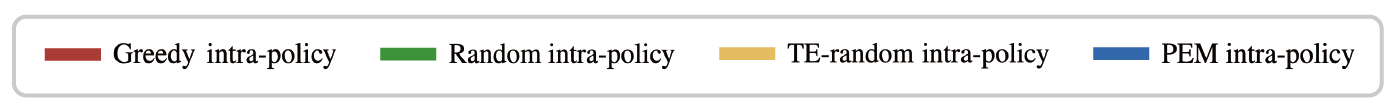}
  \end{center}
  \begin{subfigure}{0.33\linewidth}
      \includegraphics[width=.5\linewidth]{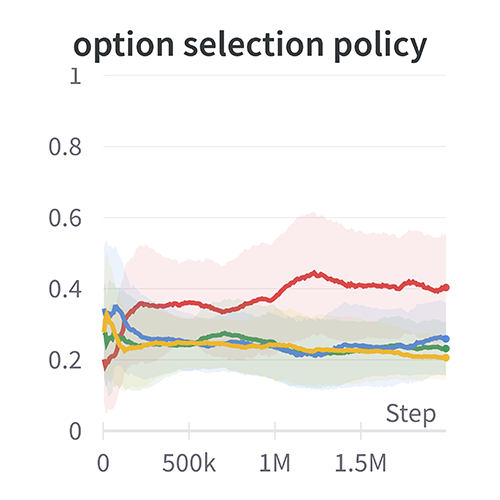}\hfill
      \includegraphics[width=.5\linewidth]{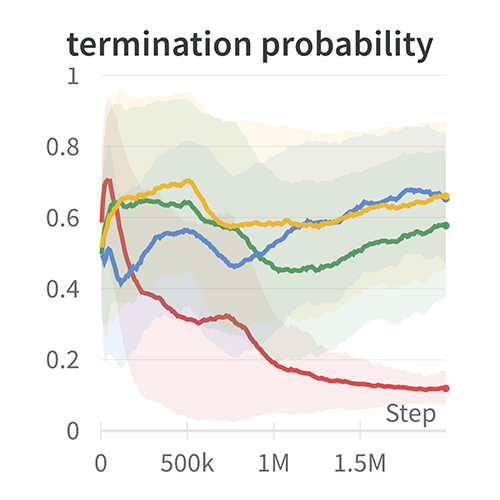}
  \caption{LavaCrossingS9N1}
  \end{subfigure}
  \begin{subfigure}{0.33\linewidth}
      \includegraphics[width=.5\linewidth]{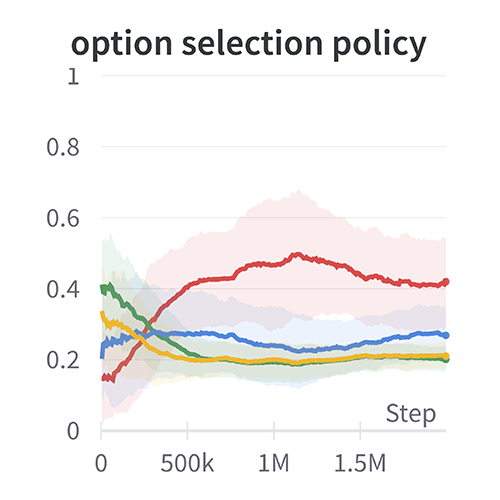}\hfill
      \includegraphics[width=.5\linewidth]{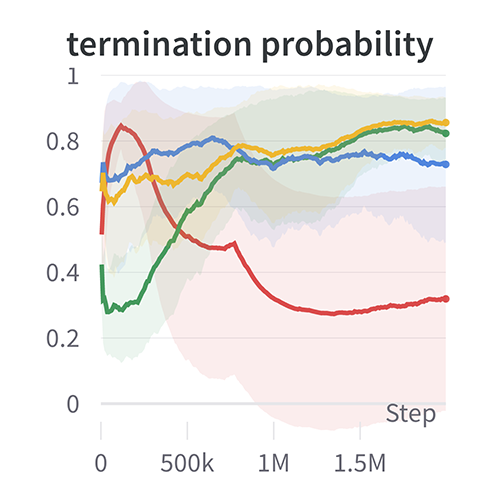}
  \caption{Fetch-8x8-N3}
  \end{subfigure}
  \begin{subfigure}{0.33\linewidth}
      \includegraphics[width=.5\linewidth]{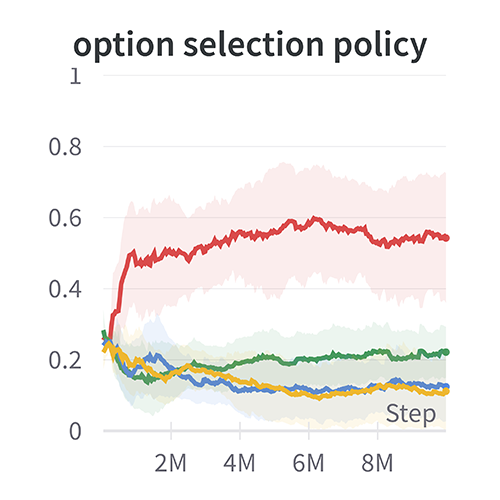}\hfill
      \includegraphics[width=.5\linewidth]{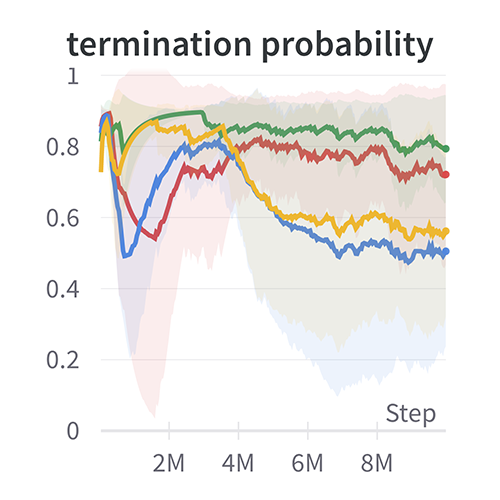}
  \caption{Enduro}
  \end{subfigure}
  \caption{Option selection policy and termination probability during training.}
  \label{fig:soft-policy-and-termination}
\end{figure*}

\begin{figure}[hbt!]
\centering
\includegraphics[width=0.5\textwidth]{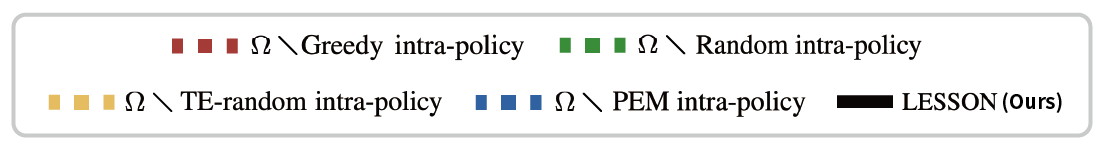}
\vspace{-1ex}
\begin{center}
\begin{tabular}{ccc}
      \hspace{-2ex}
      \includegraphics[width=0.16\textwidth]{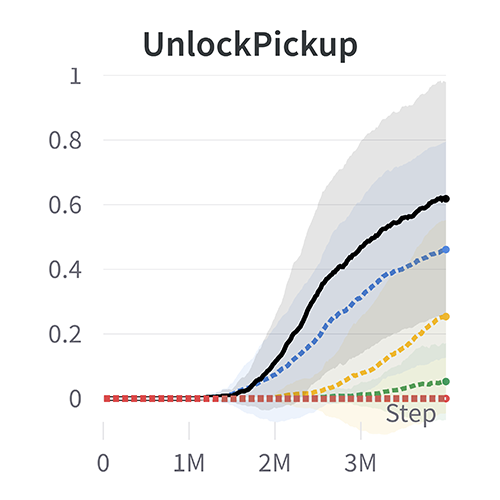} &
      \hspace{-3ex}
      \includegraphics[width=0.16\textwidth]{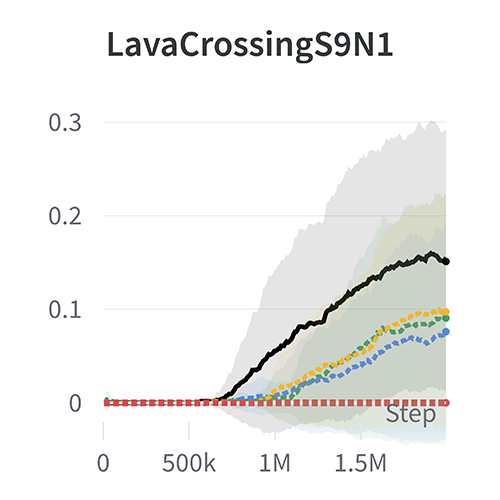} &
      \hspace{-3ex}
      \includegraphics[width=0.16\textwidth]{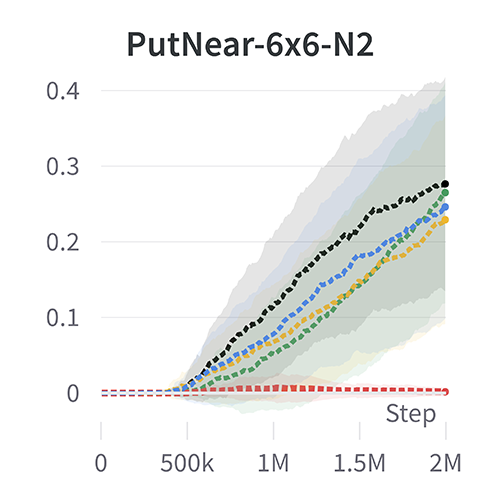} 
\end{tabular}
\end{center}
\vspace{-3ex}
\caption{Performance of LESSON that excludes one intra-policy from  $\Omega$ = $\{$greedy, random, TE-random and PEM$\}$}
\label{fig:option-combo}
\end{figure}

\subsection{Analysis} \label{sec:analysis}

\textbf{Exploration Behavior Analysis} ~~ In order to see how LESSON achieves adaptive  exploration-exploitation trade-off for better performance, we conducted an experiment on the Empty-16x16 environment. This environment is made up of a two-dimensional grid and an agent. The goal is that the agent starting from the left upper corner reaches the green box at the right lower corner as in Fig. \ref{fig:visitation} (a). 
The performance of LESSON and the baselines is shown in Fig. \ref{fig:visitation} (b), showing that LESSON outperforms others.  Fig. \ref{fig:visitation} (c) shows 
the  termination probabilities of the four intra-policies of LESSON as learning progresses. It is seen that LESSON adaptively selects the most effective intra-policy as time progresses. In the early phase of training, the termination probability of PEM intra-policy is very low compared with all others. This means that intrinsically-motivated exploration is mostly adopted in the early phase.  As time elapses, other intra-policies kick in, especially TE-random intra-policy plays a role on top of PEM intra-policy.  As time elapses further, the greedy intra-policy becomes dominant (the red line is close to zero) over other exploration intra-policies. The reason is as follows. Note that the behavior policy of LESSON tries to maximize the sum of extrinsic and intrinsic returns as seen in (\ref{eq:obj_optioncritic}). However, it is difficult to get extrinsic rewards initially and the reward is mostly intrinsic from exploration in the initial phase, but once the agent starts knowing how to reach the goal after sufficient exploration, the agent can get large extrinsic rewards from the environment.  Indeed, LESSON realizes the desired adaptive exploration-exploitation trade-off over time for a given task.  
Fig. \ref{fig:visitation} (d) shows the number of visitations for each grid point. The visitation pattern of RND looks like a quarter  circle originating from the left upper corner, whereas  
that of $\epsilon$z-greedy is a shape consisting of multiple  straight lines.  This is because RND tries to visit unvisited grid points from the past history and exploration starts from  the left upper corner, while $\epsilon$z-greedy repeats the same random action multiple times until termination. 
In contrast, the visitation pattern of LESSON covers all the state space by combining these two patterns. This behavior is well observed from the termination functions as functions of time in Fig. \ref{fig:visitation} (c), where the termination probabilities of PEM and TE-random intra-policies are small in the middle phase of learning. By efficiently using RND followed by $\epsilon$z-greedy, 
 LESSON achieves the goal faster than the baselines.

\textbf{Learning Option-Critic Model} ~~ Fig. \ref{fig:soft-policy-and-termination} shows  the behavior of the (soft) option selection policy $\pi_\Omega$  and the termination probabilities together for several tasks. Note that the option selection policy   and the termination probabilities together determine the frequency of use of each intra-policy. 
It is seen that the selection probability of the greedy intra-policy tends to increase during training for all tasks, as expected.   In LavaCrossing and Fetch, where the $\epsilon$z-greedy performs poorly (see Fig. \ref{fig:OverallResults} (b) and (e)), the termination probability of TE-random  intra-policy (comprising $\epsilon$z-greedy) is higher than those of other exploration strategies together  with a low selection probability for TE-random  intra-policy, as seen in Fig. \ref{fig:soft-policy-and-termination} (a) and (b). In these tasks, LESSON exploits the integration of random and PEM intra-policies (i.e., $\epsilon$r-greedy and RND).  Note that EWC performs worse than $\epsilon$r-greedy since it equally exploits all exploration strategies including ineffective $\epsilon$z-greedy.    In Enduro, on the other hand,
the termination probability of random intra-policy (comprising $\epsilon$r-greedy) is higher than others. In this task,  LESSON exploits both TE-random and PEM intra-policies (i.e., $\epsilon$z-greedy and RND). Indeed, LESSON successfully learns to use suitable exploration strategies depending on the task.

\begin{figure}[t!]
\centering
\includegraphics[width=0.35\textwidth]{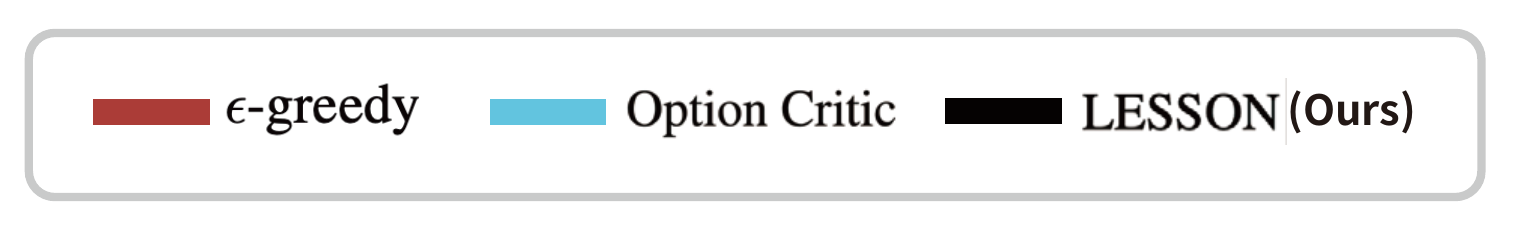}
\vspace{-1ex}
\begin{center}
\begin{tabular}{ccc}
      \hspace{-2ex}
      \includegraphics[width=0.16\textwidth]{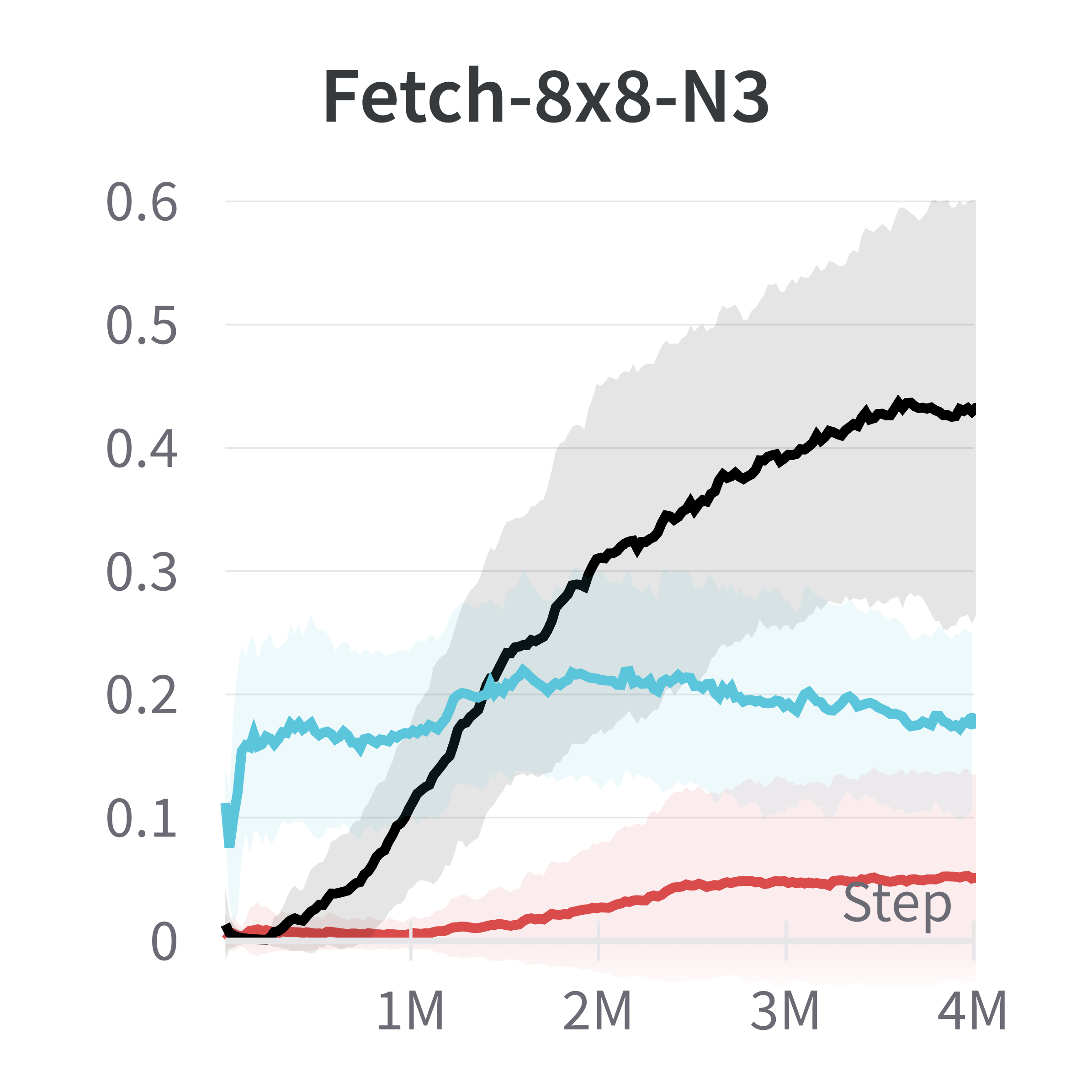} &
      \hspace{-3ex}
      \includegraphics[width=0.16\textwidth]{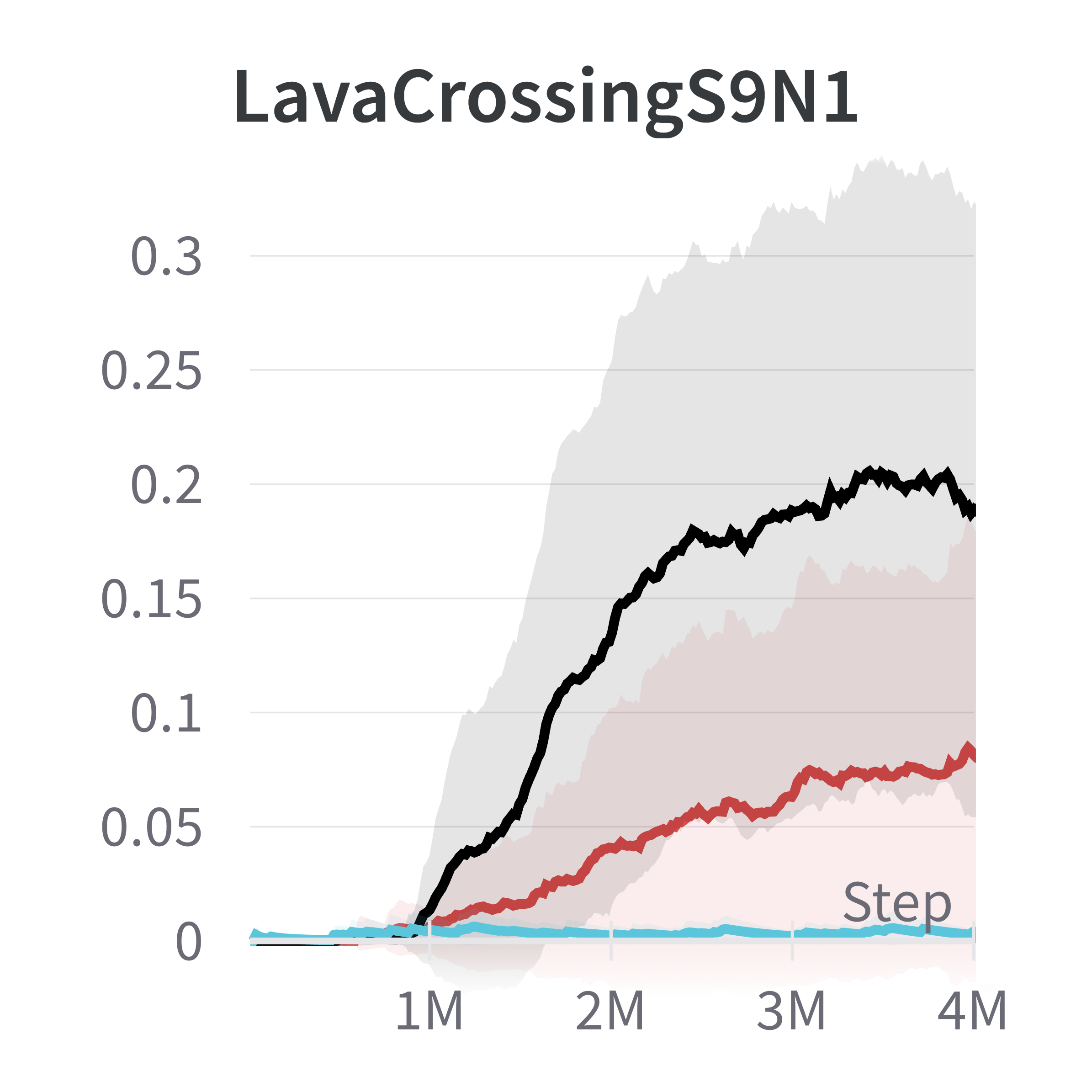} &
      \hspace{-3ex}
      \includegraphics[width=0.16\textwidth]{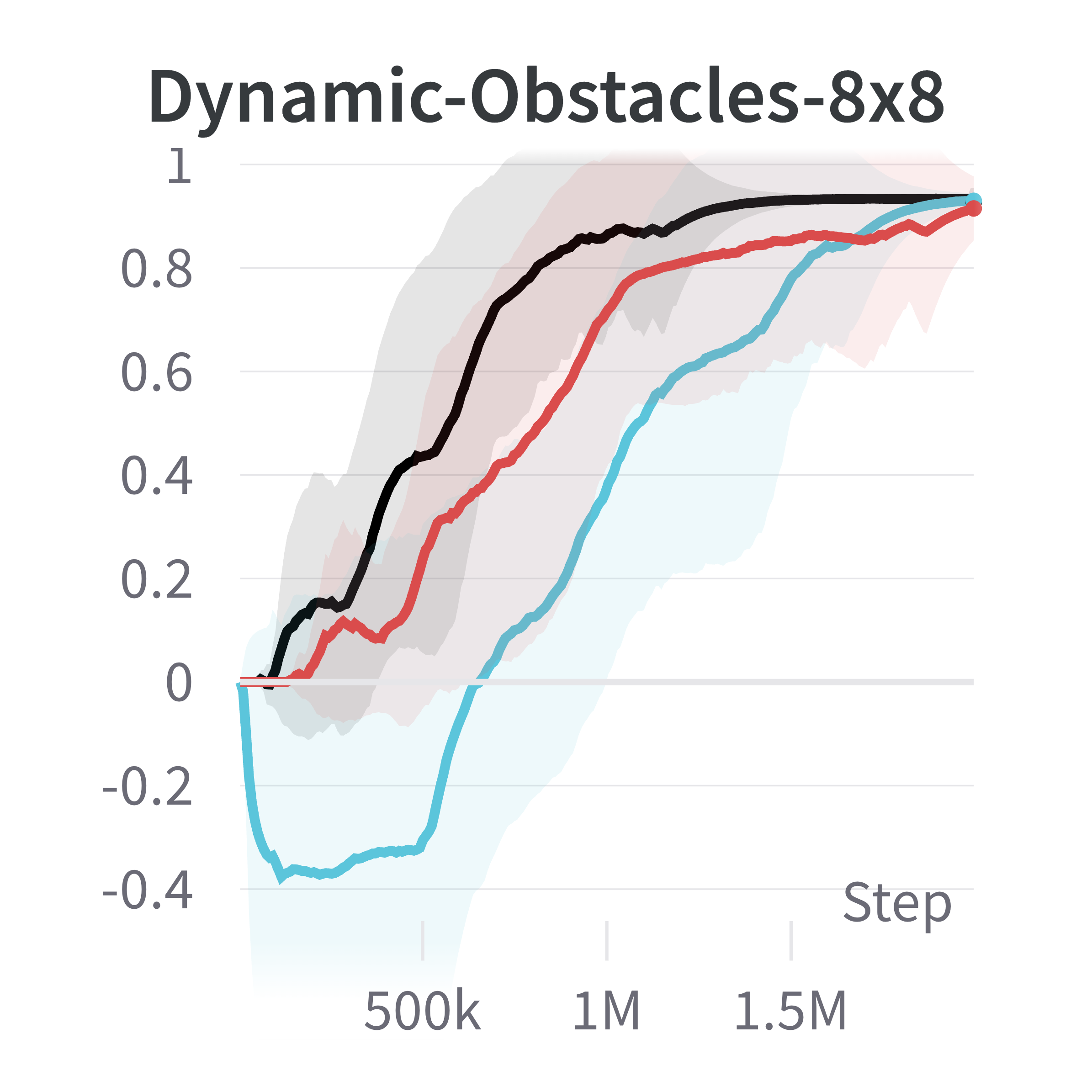} 
\end{tabular}
\end{center}
\vspace{-3ex}
\caption{Performance comparison of LESSON against $\epsilon$-greedy and option-critic}
\label{fig:option_scratch}
\end{figure}

\textbf{Ablation Study: Learning Intra-policies from Scratch} ~~
We used a set of pre-defined intra-policies as options for LESSON rather than  learning options from scratch. To verify the  effectiveness of this pre-defined option approach, we conducted an experiment that learns options from scratch  with the objective function of LESSON. Such an approach of learning of options from scratch is originally considered in the option critic architecture \cite{bacon2017option}. This learning architecture selects high-level options and actions based on each option without pre-defining options, and only pre-defines the number of options as a hyperparameter. We set the number of options as 4 to be equal to that used for LESSON in this paper, and trained the option critic by using an intrinsic reward similar to that of LESSON.
Fig. \ref{fig:option_scratch} shows the result of this experiment, showing that LESSON outperforms the option-critic architecture learning options from scratch. The  poor performance of this approach  seems to result from the difficulty associated with comprehending and acquiring temporal abstraction solely through rewards (with exploration bonus) in scenarios with sparse settings like MiniGrid.

\textbf{Ablation Study: Impact of Each Intra-policy} ~~ 
We investigated the performance of LESSON by eliminating one intra-policy from $\Omega$ = $\{$greedy, random, TE-random and PEM$\}$. The result is shown in Fig. \ref{fig:option-combo}. It is seen that LESSON is unable to learn without the inclusion of the greedy policy. Thus, the inclusion of the greedy policy within the set of intra-policies, which is one of the main ideas of LESSON, is necessary to realize exploitation in addition to exploration for the behavior policy.  It is also seen that performance tends to be degraded most if the most effective exploration strategy for each environment is excluded. 

\begin{figure}[t!]
\centering
\includegraphics[width=0.4\textwidth]{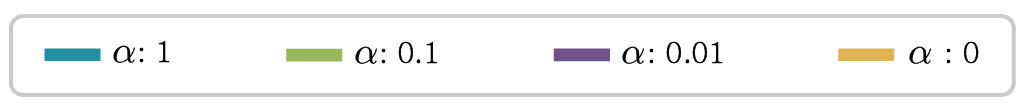}
\vspace{-1ex}
\begin{center}
\begin{tabular}{ccc}
      \hspace{-2ex}
      \includegraphics[width=0.16\textwidth]{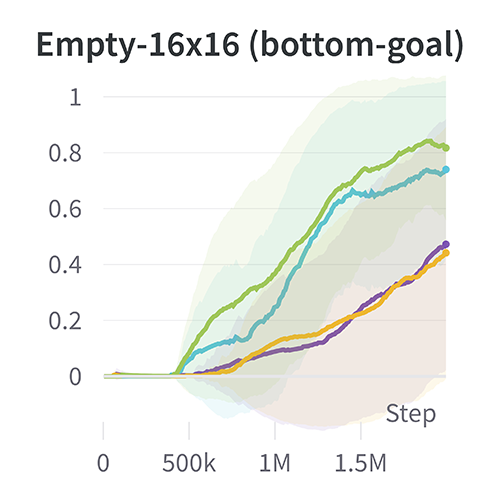} &
      \hspace{-3ex}
      \includegraphics[width=0.16\textwidth]{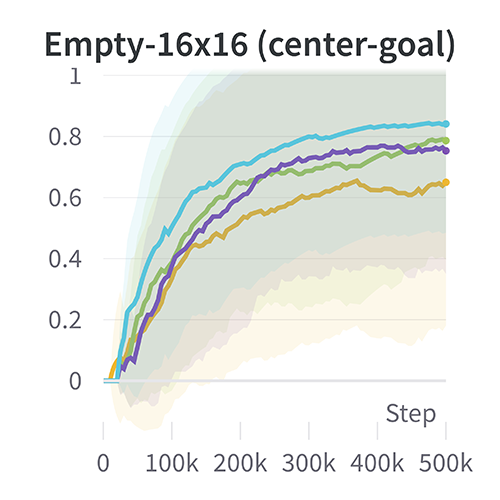} &
      \hspace{-3ex}
      \includegraphics[width=0.16\textwidth]{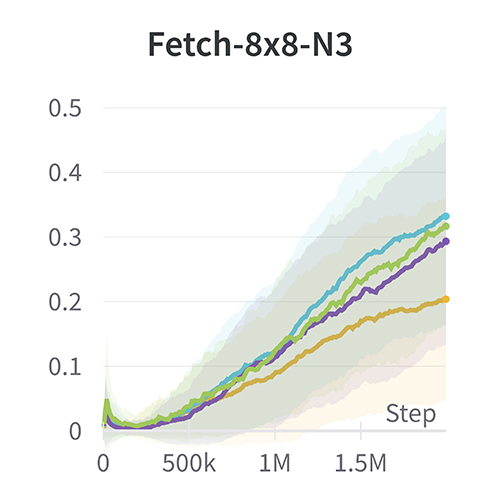} 
\end{tabular}
\end{center}
\vspace{-2ex}
\caption{Performance of LESSON with respect to the intrinsic reward coefficient $\alpha$.}
\label{fig:ablation_alpha}
\end{figure}

\textbf{Ablation Study: Impact of $\alpha$} ~~ We investigated the impact of $\alpha$ determining the ratio between extrinsic and intrinsic rewards in the objective function \eqref{eq:obj_optioncritic} for the behavior policy. Fig. \ref{fig:ablation_alpha} shows the performance of LESSON w.r.t. $\alpha$. The coefficient $\alpha$ should be set properly to obtain the desired exploration-exploitation trade-off over time.  If $\alpha$ is too small, then exploration is not performed well and it takes a long time to learn the task purely based on extrinsic rewards in sparse reward cases.
On the other hand, if $\alpha$ is too large and hence the intrinsic reward portion is too large compared with the extrinsic reward portion, then the behavior policy will try exploration persistently for large intrinsic rewards even if it knows how to solve the task and get extrinsic rewards. However, due to our use of the history-based prediction-error intrinsic reward, when all state-action pairs are visited sufficiently many times, the estimation error for all state-action combinations  becomes  small, the corresponding intrinsic reward becomes small, and hence extrinsic reward becomes dominant eventually. Hence, too large $\alpha$ also delays the learning. Recall that we always have a separate greedy target policy learning from samples from the behavior policy, but the drawn samples affect the learning speed and performance of the target policy. 

\section{Conclusion}

We have proposed LESSON to automatically choose  an appropriate  exploration strategy from a given set  to realize an effective exploration-exploitation trade-off over time. LESSON is based on an  
 option-critic model of which  intra-policies consist of the greedy policy and  a set of diverse exploration strategies. We have designed the option-critic model judiciously  by defining relevant objectives and action value functions to realize adaptive selection of exploitation or exploration strategies.   
Although LESSON has  the increased complexity compared to existing  other exploration methods and more learnable parameters,  
 LESSON eliminates the necessity of human trial of multiple exploration strategies for each given task, and numerical results show its effectiveness. In this paper, we have demonstrated the effectiveness of learning to integrate multiple exploration strategies via an option framework  primarily based on deep Q-learning. However, such exploration integration learning is not  restricted to  Q-learning. We expect that application of the idea of LESSON to other advanced RL algorithms beyond Q-learning 
 can  enhance their learning speed and/or  performance even further.

\section*{Acknowledgements}

This work was supported  by Center for Applied Research in Artificial Intelligence (CARAI) Grant funded by Defense Acquisition Program Administration (DAPA) and Agency for Defense Development (ADD) of Repulic of Korea (UD230017TD).

\newpage

\bibliography{references}
\bibliographystyle{icml2023}

\newpage
\appendix
\onecolumn
\section{Environment Specifications}  \label{sec:appx-environments}

\subsection{MiniGrid}

MiniGrid \cite{gym_MiniGrid} is a collection of 2D grid-world goal-based environments.
The agent receives a sparse reward $R_1$ with a small decrement for each interaction step. In this paper, we set this reward as $R_1=10$.  
 The considered MiniGrid environments are briefly explained below:

\begin{figure*}[hbt!]
\centering
\begin{subfigure}{0.14\textwidth}
    \centering
    \includegraphics[width=\textwidth]{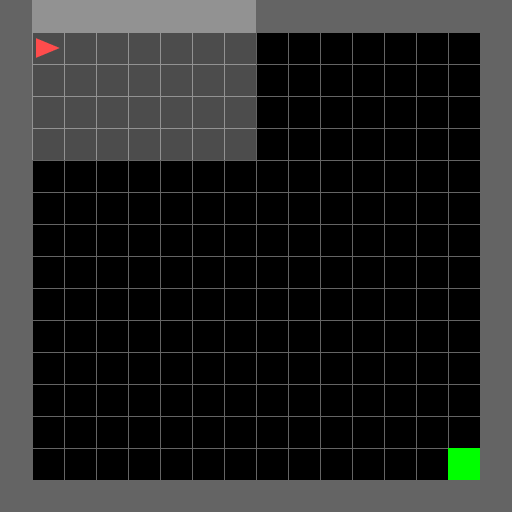}
    \caption{Empty-16x16 \break}
\end{subfigure}
\begin{subfigure}{0.14\textwidth}
    \centering
    \includegraphics[width=\textwidth]{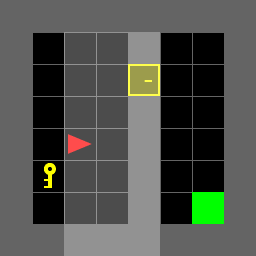}
    \caption{DoorKey-8x8 \break}
\end{subfigure}
\begin{subfigure}{0.13\textwidth}
    \centering
    \includegraphics[width=\textwidth]{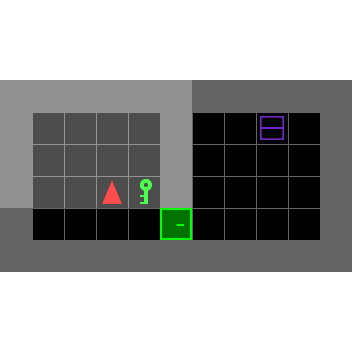}
    \caption{UnlockPickup \break}
\end{subfigure}
\begin{subfigure}{0.13\textwidth}
    \centering
    \includegraphics[width=\textwidth]{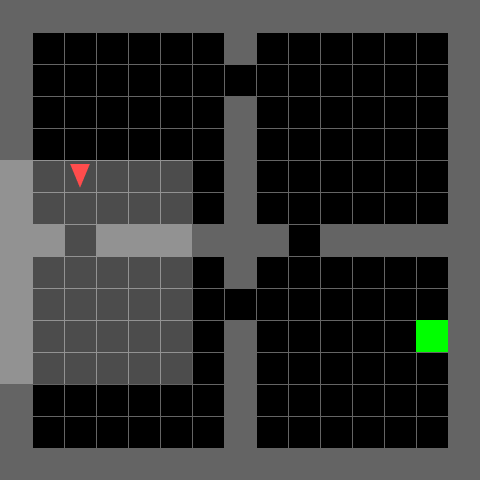}
    \caption{FourRooms \break}
\end{subfigure}
\begin{subfigure}{0.13\textwidth}
    \centering
    \includegraphics[width=\textwidth]{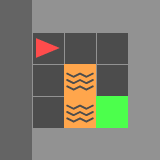}
    \caption{LavaGapS5 \break}
\end{subfigure}
\begin{subfigure}{0.13\textwidth}
    \centering
    \includegraphics[width=\textwidth]{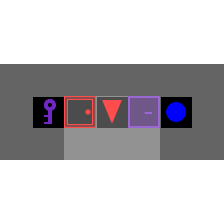}
    \caption{KeyCorridorS3R1}
\end{subfigure}
\begin{subfigure}{0.13\textwidth}
    \centering
    \includegraphics[width=\textwidth]{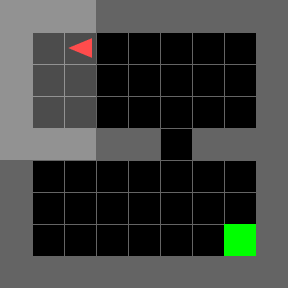}
    \caption{SimpleCrossingS9N1}
\end{subfigure}
\begin{subfigure}{0.13\textwidth}
    \centering
    \includegraphics[width=\textwidth]{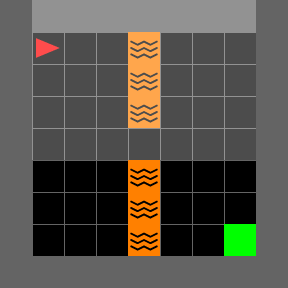}
    \caption{LavaCrossingS9N1}
\end{subfigure}
\begin{subfigure}{0.13\textwidth}
    \centering
    \includegraphics[width=\textwidth]{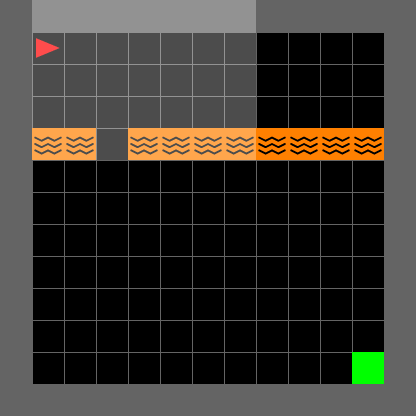}
    \caption{LavaCrossingS13N1}
\end{subfigure}
\begin{subfigure}{0.13\textwidth}
    \centering
    \includegraphics[width=\textwidth]{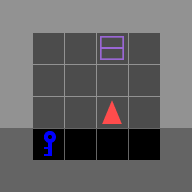}
    \caption{PutNear-6x6-N2 }
\end{subfigure}
\begin{subfigure}{0.13\textwidth}
    \centering
    \includegraphics[width=\textwidth]{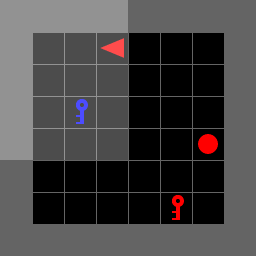}
    \caption{Fetch-8x8-N3 \break} 
\end{subfigure}
\begin{subfigure}{0.13\textwidth}
    \centering
    \includegraphics[width=\textwidth]{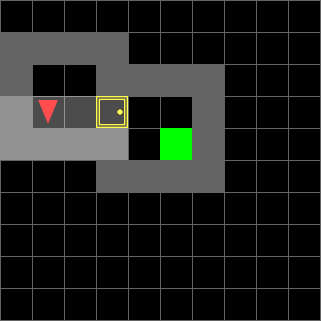}
    \caption{MultiRoom-N2-S4}
\end{subfigure}
\begin{subfigure}{0.13\textwidth}
    \centering
    \includegraphics[width=\textwidth]{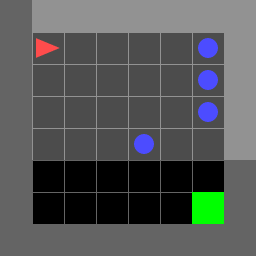}
    \caption{Dynamic-Obstacles-8x8}
\end{subfigure}
\begin{subfigure}{0.13\textwidth}
    \centering
    \includegraphics[width=\textwidth]{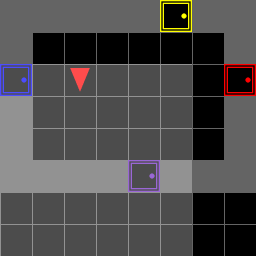}
    \caption{GoToDoor-8x8}
\end{subfigure}
\caption{MiniGrid Environments used in the experiments. These figures were obtained by rendering the MiniGrid environment while training \cite{gym_MiniGrid} . }
\label{fig:MiniGrid-envs}
\end{figure*}

\textbf{Empty-16x16} The agent is initially located at the top-left of an empty room, and the goal is to reach the green goal square. 

\textbf{DoorKey-8x8} The agent should pick up a key to unlock a door, and then should navigate to the green goal square.

\textbf{UnlockPickup} The agent should acquire a box that is located in a different room, which can only be accessed through  a locked door.

\textbf{FourRooms} Each of the agent and the green goal is  randomly initialized in one of the four rooms.  Then, the agent should search the rooms to reach the green goal.

\textbf{LavaGapS5} The agent should reach the green goal  located in the opposite corner of the room. In order to reach the  goal, the agent should  pass through a narrow opening in a vertical strip of deadly lava. If the agent touches the lava, the episode is terminated. 

\textbf{KeyCorridorS3R1} The agent should search the map to find the key that  is hidden in another room, and then should pick up the object that is located behind the locked door.

\textbf{SimpleCrossingS9N1, LavaCrossingS9N1, LavaCrossingS13N1} The agent should reach a goal while avoiding obstacles that randomly block single row or column with one square opening within the grid in each environment. The difference between SimpleCrossing tasks and LavaCrossing tasks is that collision with an obstacle results in the failure of the episode in LavaCrossing tasks, whereas the episode continues even if an  obstacle is encountered in SimpleCrossing tasks.

\textbf{PutNear-6x6-N2} The agent receives instructions in the form of a textual string (mission) such as "picking up the object" and then "placing it next to another object". The agent receives a reward when accurately executing the provided instruction.

\textbf{Fetch-8x8-N3} A textual string (mission) as part of its observation indicating which object to pick up is provided. The environment contains various objects of different types and colors, and an incorrect selection of an object results in the termination of the episode with no reward.

\textbf{MultiRoom-N2-S4} The environment consists of two rooms and the agent should reach the green goal, which is located in the next room. In order to obtain access to the next room, the agent needs to open the doors.

\textbf{Dynamic-Obstacles-8x8} The agent should  reach the green goal square while avoiding moving obstacles. If the agent collides with an obstacle, the agent receives a large penalty and the episode is terminated.

\textbf{GoToDoor-8x8} The environment consists of a room with four doors, one on each wall, and a textual string (mission) as input which indicates the target door to be reached (e.g. "go to the red door"). When successfully reaching the correct door as indicated by the mission string, the agent receives a positive reward.

\subsection{Atari 2600} 

\begin{figure}[hbt!]
    \centering
    \includegraphics[width=0.85\linewidth]{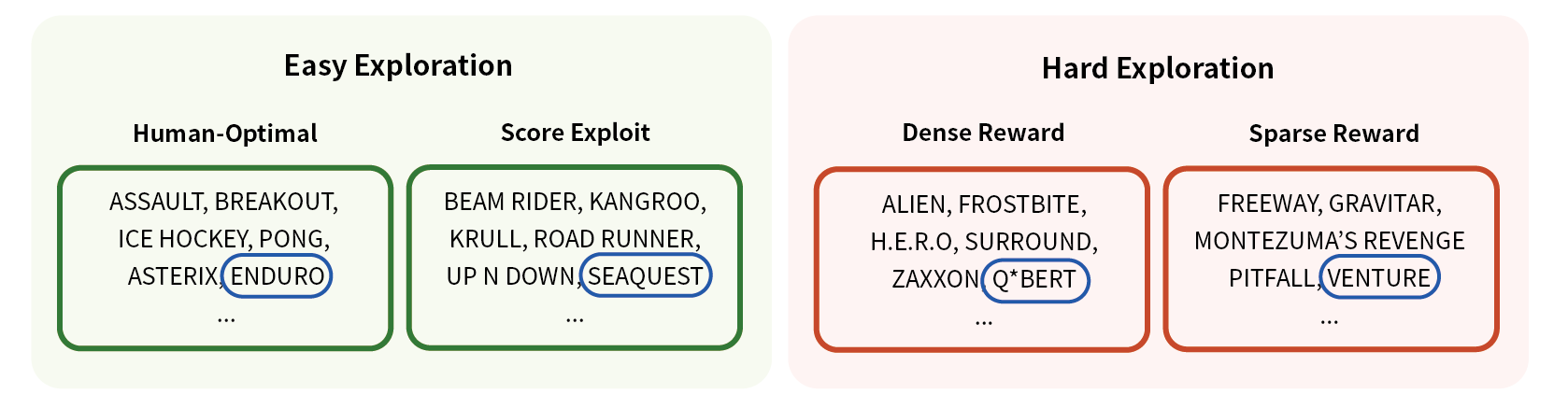}
    \caption{A taxonomy for experiments of Atari 2600 games based on the level of the exploration difficulty from \citep{bellemare2016unifying}. The environments we experimented are  marked with blue ovals.}
    \label{fig:atari-taxonomy}
\end{figure}

We conducted experiments on the Arcade Learning Environment (ALE, \citet{bellemare2012ale}) at various difficulty levels and reward settings. ALE offers a comprehensive interface to a wide range of Atari 2600 game environments, which are known to be challenging even for human players. These games require long-term credit assignment and difficult exploration \cite{badia2020agent57}. Atari games  are classified into four groups, based on the various characteristics and difficulty, e.g.,  whether local exploration methods such as $\epsilon$-greedy are sufficient or not \citet{bellemare2016unifying}. The classification is shown in  Fig. \ref{fig:atari-taxonomy}. In order to include  a diverse range of experiments, we conducted experiments with  one environment from each of these four groups.

\begin{figure*}[hbt!]
\centering
\begin{subfigure}{0.15\textwidth}
    \centering
    \includegraphics[width=\textwidth, height=1.3in]{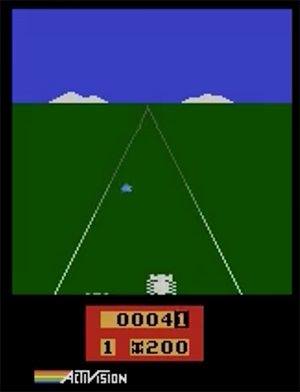}
    \caption{Enduro}
\end{subfigure}
\begin{subfigure}{0.15\textwidth}
    \centering
    \includegraphics[width=\textwidth, height=1.3in]{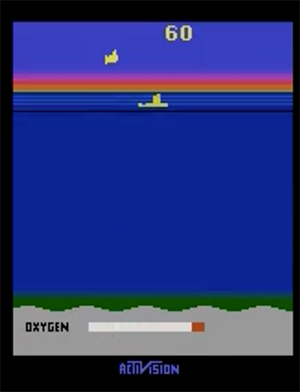}
    \caption{Seaquest}
\end{subfigure}
\begin{subfigure}{0.15\textwidth}
    \centering
    \includegraphics[width=\textwidth, height=1.3in]{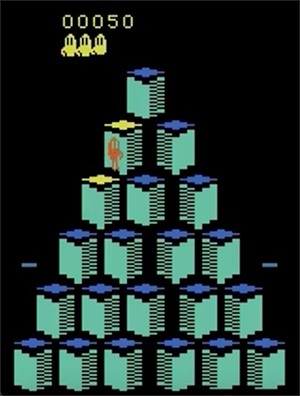}
    \caption{Qbert}
\end{subfigure}
\begin{subfigure}{0.15\textwidth}
    \centering
    \includegraphics[width=\textwidth, height=1.3in]{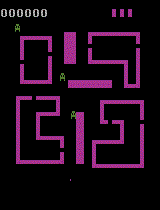}
    \caption{Venture}
\end{subfigure}
\caption{Atari 2600 Environments used in the experiments. The source of these figures is \url{https://www.gymlibrary.dev/environments/atari}}
\label{fig:atari-envs}
\end{figure*}

\textbf{Enduro} In the National Enduro endurance race, the goal is to overtake a specified number of cars each day in order to remain in the race. On the first day, the player should  overtake 200 cars, and on subsequent days, the number increases to be 300 cars per day. The game ends if the player fails in meeting the required number of overtakes for a given day.

\textbf{Seaquest} In this game, the player controls a submarine and should  retrieve divers while avoiding and attacking enemy subs and sharks. Points are awarded based on performance. The player  can earn more as their score increases but can only have six on screen at a time. When colliding with anything other than a diver,   the submarine explodes, and there is a limited supply of oxygen. If the player fails in surfacing  in time or has less than six divers, they lose one diver.

\textbf{Qbert} In this game, the player assumes the role of Q*bert of which  objective is to alter the color of all the cubes on a pyramid to match the designated destination color. The player should 
accomplish this task by hopping on each cube of the pyramid in sequence, while avoiding hostile creatures  in the pyramid.

\textbf{Venture} The goal of this game is to successfully navigate through a dungeon, collecting treasure in every chamber while eliminating any monsters that may be present. The player should make a careful strategy  and act to achieve this goal.

\newpage

\section{Implementation Details} \label{sec:appx-implementation-details}

In this section, we provide the training details including hyperparameters. 

\textbf{Baselines.} ~~The three baselines including $\epsilon$r-greedy, $\epsilon$z-greedy, and EWC determine the duration of random or fixed action. The duration is sampled from a zeta distribution, represented by $\zeta(n) \propto n^{-\mu}$, with $\mu = 2$ as in \citep{dabney2020temporally}.

\subsection{Architecture of Neural Networks}

\begin{figure}[hbt!]
    \centering
    \includegraphics[width=0.8\linewidth]{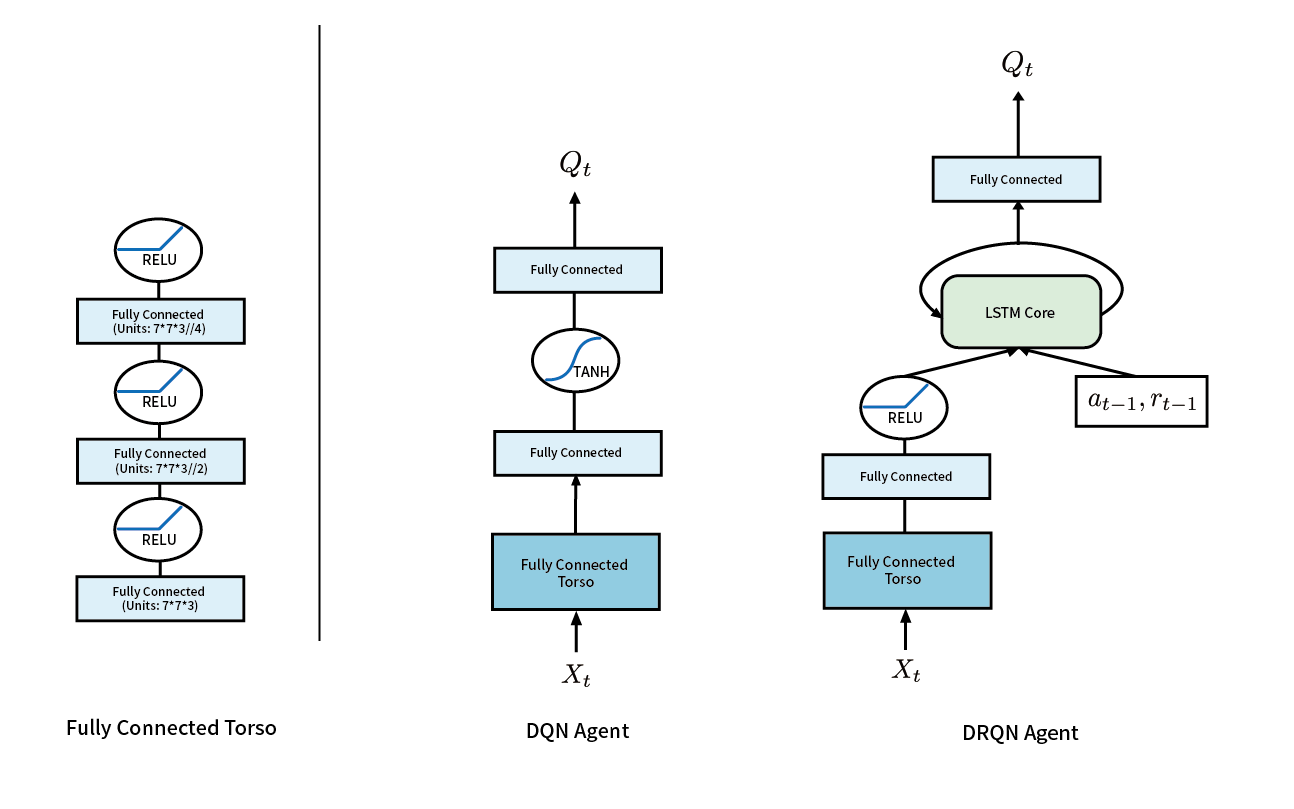}
    \caption{MiniGrid Agent Architecture}
    \label{fig:MiniGrid-networks}
\end{figure}

\textbf{MiniGrid.} ~~In the considered MiniGrid tasks except for the FourRooms task, the proposed algorithm and the considered baselines were implemented on the top of  Deep Q-Network (DQN). For the FourRooms task, we adopted  Deep Recurrent Q-Network (DRQN) as it is capable of integrating information across frames to detect relevant information, which is particularly helpful in environments like FourRooms in wich  the ability to gather information from the previously visited rooms is beneficial. The designed neural network architectures are shown in Fig. \ref{fig:MiniGrid-networks}.

\textbf{Atari.} ~~ In this case, all the algorithms were  implemented on top of DQN, which is modeled based on the CNNPolicy. The detail parameters regarding the CNNPolicy were determined based on the default architecture in stable-baselines3 \citep{stable-baselines3}.

\subsection{Hyper-parameters}

\textbf{MiniGrid.} ~~The value of $\epsilon$ was decreased linearly from 0.9 to 0.05 over $10^5$
steps. The target update period was 1000, the replay buffer size was $5 \times 10^5$, and the mini-batch size was 256 trajectories for DQN and 1 episode for DRQN. In all the environments, we employed the RMSProp optimizer with a learning rate of 0.0001 for the learning agent, and the Adam optimizer for the RND predictor network. The maximum number of steps in one episode was 40 for MultiRoom-N2-S4, 60 for PutNear-6x6-N2, and 100 for the remaining tasks.

All the networks including the option-critic model and the Q-functions regarding extrinsic reward and intrinsic reward were trained at intervals of 10 time steps in the considered MiniGrid tasks except for the PutNear-6x6 environment. For the PutNear-6x6 environment, the option-critic model is trained at intervals of 4 time steps in order to accelerate the change in termination probabilities.

During the learning process of RND and LESSON agents, the intrinsic reward coefficient $\alpha$ was tuned among the values (0.001, 0.01, 0.1, 1) and for LESSON agents, the temperature parameter $\tau$ in option selection policy was tuned among the values (0.02, 0.2). The used $\alpha$ and $\tau$ value are provided in Table \ref{table:hyper-parameters-MiniGrid}.

\begin{table}[hbt!]
    \centering
    \caption{Hyper-parameters values used in MiniGrid}
    \label{table:hyper-parameters-MiniGrid}
    \begin{subtable}{\textwidth}
        \centering
        \begin{tabular}{c|cc}
        \noalign{\smallskip}\noalign{\smallskip}\hline\hline
        Hyperparameter & Intrinsic reward coefficient $\alpha$ & Temperature parameter $\tau$ \\
        \hline
        Empty-16x16 (bottom-goal)  & 0.1 & 0.02 \\
        Empty-16x16 (center-goal)  & 1 & 0.02 \\
        DoorKey-8x8  & 0.01  & 0.02\\
        UnlockPickup & 0.1 & 0.2 \\
        FourRooms & 0.1 & 0.02 \\
        LavaGapS5  & 0.1 & 0.2 \\
        KeyCorridorS3R1  & 0.01 & 0.2 \\
        SimpleCrossingS9N1  & 0.001 & 0.02 \\
        LavaCrossingS9N1  & 0.01 & 0.02 \\
        LavaCrossingS13N1  & 0.01 & 0.02 \\
        PutNear-6x6-N2  & 0.1 & 0.02 \\
        Fetch-8x8-N3  & 1 & 0.2 \\
        MultiRoom-N2-S4  & 0.1 & 0.2 \\
        Dynamic-Obstacles-8x8  & 0.01 & 0.02 \\
        GoToDoor-8x8  & 0.1 & 0.2 \\
        DoorKey-8x8  & 0.01 & 0.02 \\
        
        \hline
        \hline
        \end{tabular}
        \caption{Hyper-parameters used in LESSON agents} 
    \end{subtable}
    \begin{subtable}{0.5\textwidth}
        \begin{tabular}{c|c}
        \noalign{\smallskip}\noalign{\smallskip}\hline\hline
        Hyperparameter & Intrinsic reward coefficient $\alpha$ \\
        \hline
        Empty-16x16 (bottom-goal)  & 0.01  \\
        Empty-16x16 (center-goal)  & 0.01 \\
        DoorKey-8x8  & 0.01  \\
        UnlockPickup & 0.1  \\
        FourRooms & 0.1  \\
        LavaGapS5  & 0.1  \\
        KeyCorridorS3R1  & 0.01  \\
        SimpleCrossingS9N1  & 0.001  \\
        LavaCrossingS9N1  & 0.01  \\
        LavaCrossingS13N1  & 0.01  \\
        PutNear-6x6-N2  & 0.1  \\
        Fetch-8x8-N3  & 1  \\
        MultiRoom-N2-S4  & 0.1  \\
        Dynamic-Obstacles-8x8  & 0.1  \\
        GoToDoor-8x8  & 1  \\
        \hline
        \hline
        \end{tabular}
        \caption{Hyper-parameters used in RND agents} 
    \end{subtable}
\end{table}

\textbf{Atari.} ~~
The value of $\epsilon$ was decreased linearly from 1 to 0.1 over $10^6$ time steps. Most of the hyperparameter setting was  based on the default setting of stable-baselines3 \citep{stable-baselines3}. However,  a few modifications were made. For example, the training frequency was changed to once per step, the start time-step of training was set to $10^4$, and the replay buffer size was set to $10^6$. In addition, all networks were updated at each time step.

During the learning process of RND and LESSON agents, the intrinsic reward coefficient $\alpha$ was tuned among the values (0.01, 0.001), and for LESSON agents, the temperature parameter $\tau$ in option selection policy was tuned among the values (0.01, 0.02). The specific values of used $\alpha$ and $\tau$ are provided in Table \tcb{\ref{table:hyper-parameters-Atari}}. In the Atari environments, the same value of $\alpha$ was used for both the RND and LESSON agents.

\begin{table}[hbt!]
    \centering
    \caption{Hyper-parameters used in Atari 2600}
    \label{table:hyper-parameters-Atari}
    \begin{subtable}{\textwidth}
        \centering
        \begin{tabular}{c|cc}
        \noalign{\smallskip}\noalign{\smallskip}\hline\hline
        Hyperparameter & Intrinsic reward coefficient $\alpha$& Temperature parameter $\tau$ \\
        \hline
        Enduro & 0.01 & 0.02 \\
        Seaquest & 0.01 & 0.01 \\
        Qbert  & 0.001 & 0.02 \\
        Venture  & 0.001 & 0.02 \\
        \hline
        \hline
        \end{tabular}
    \end{subtable}
\end{table}

\subsection{Intrinsic reward} \label{appx-rnd-details}
The intrinsic reward was generated by the prediction error  of the RND network, as described in \citep{burda2018exploration}. The additional networks, referred to as "predictor" and "target", were implemented as sequential layers of linear units with  final output size of 64. 
 The training of the RND predictor network was  performed concurrently with that of the main agent network, utilizing the same replay batches. To ensure stability in the training process, we normalized the intrinsic reward to a zero mean Gaussian distribution by using the running mean and standard deviation in a similar way to that mentioned in \citep{burda2018exploration}.

\newpage

\section{Experimental Results} \label{sec:appx-experimental-results}

\begin{figure*}[hbt!]
\begin{center}
\includegraphics[width=0.7\textwidth]{figure/label/label.pdf}
\begin{tabular}{cccc}
      \includegraphics[width=0.23\textwidth]{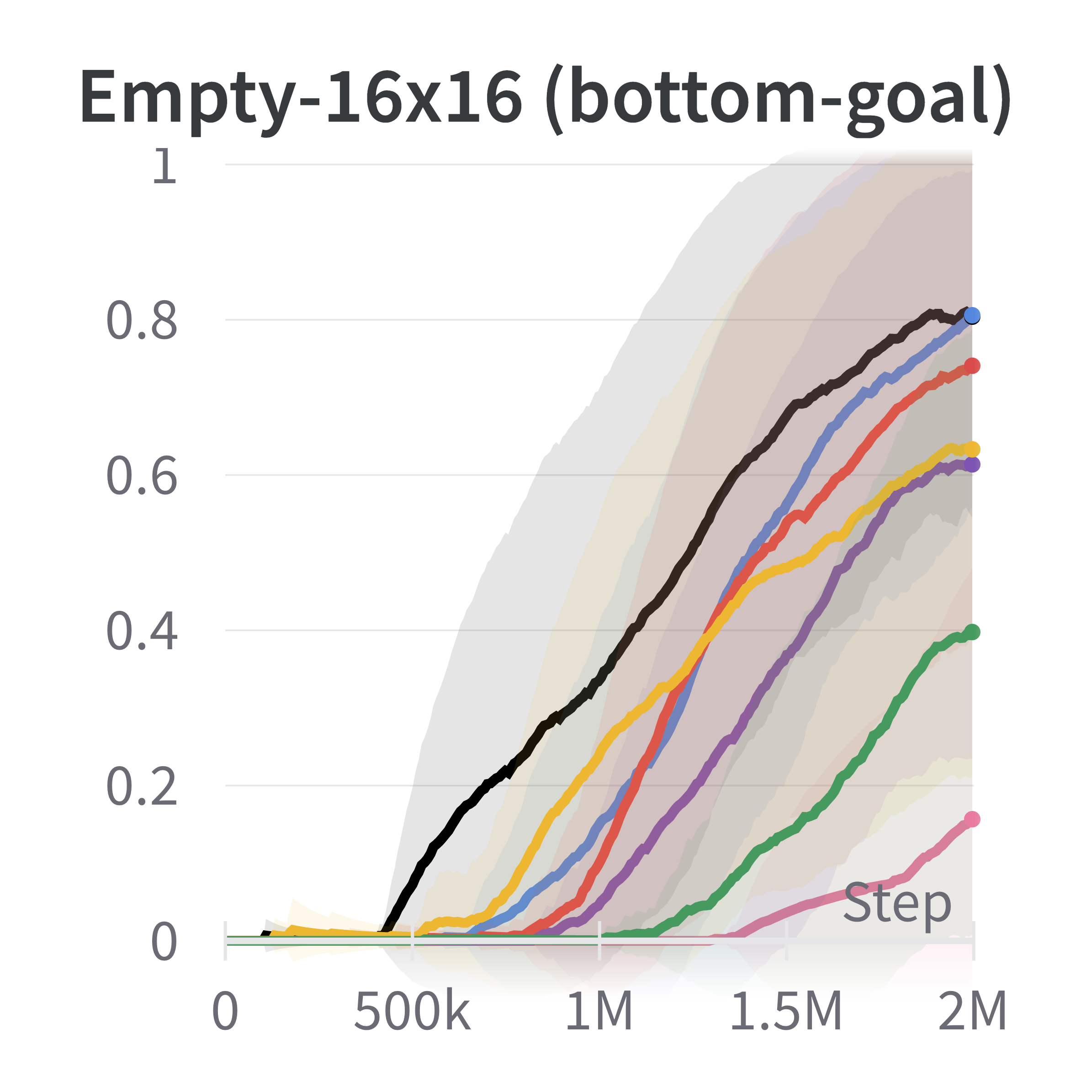} &
      \includegraphics[width=0.23\textwidth]{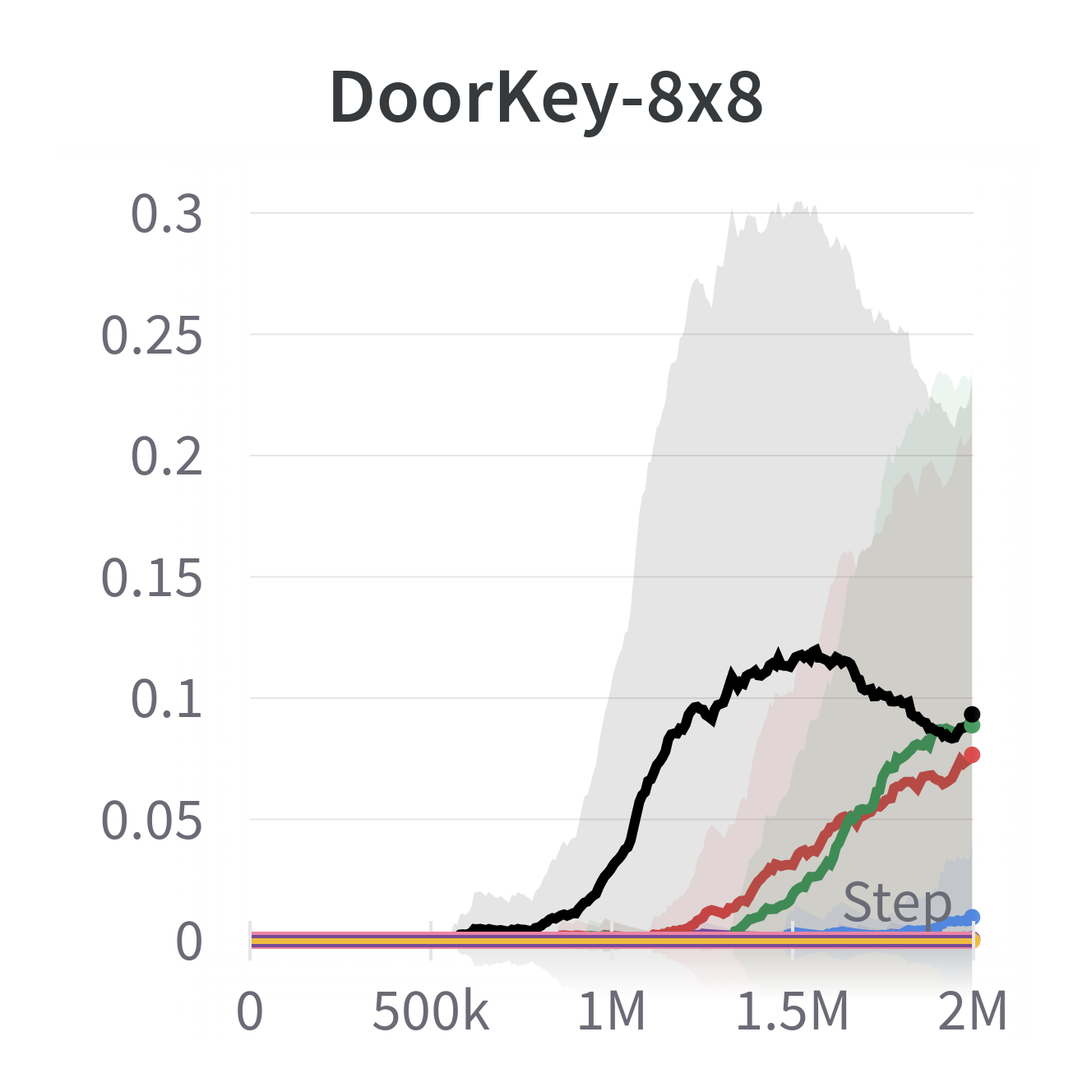} &
      \includegraphics[width=0.23\textwidth]{figure/long-unlockpickup.png} &
      \includegraphics[width=0.23\textwidth]{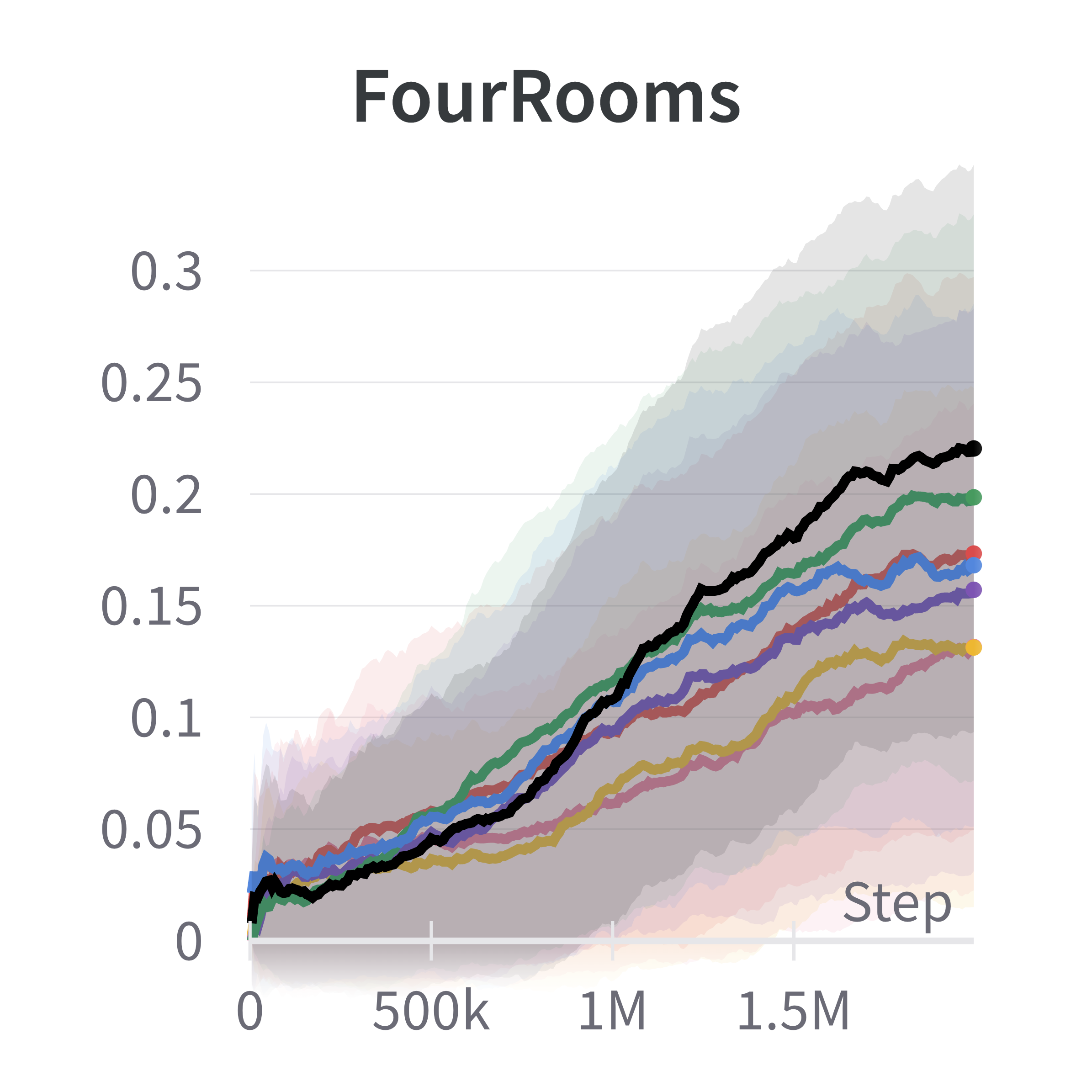} \\      
      \includegraphics[width=0.23\textwidth]{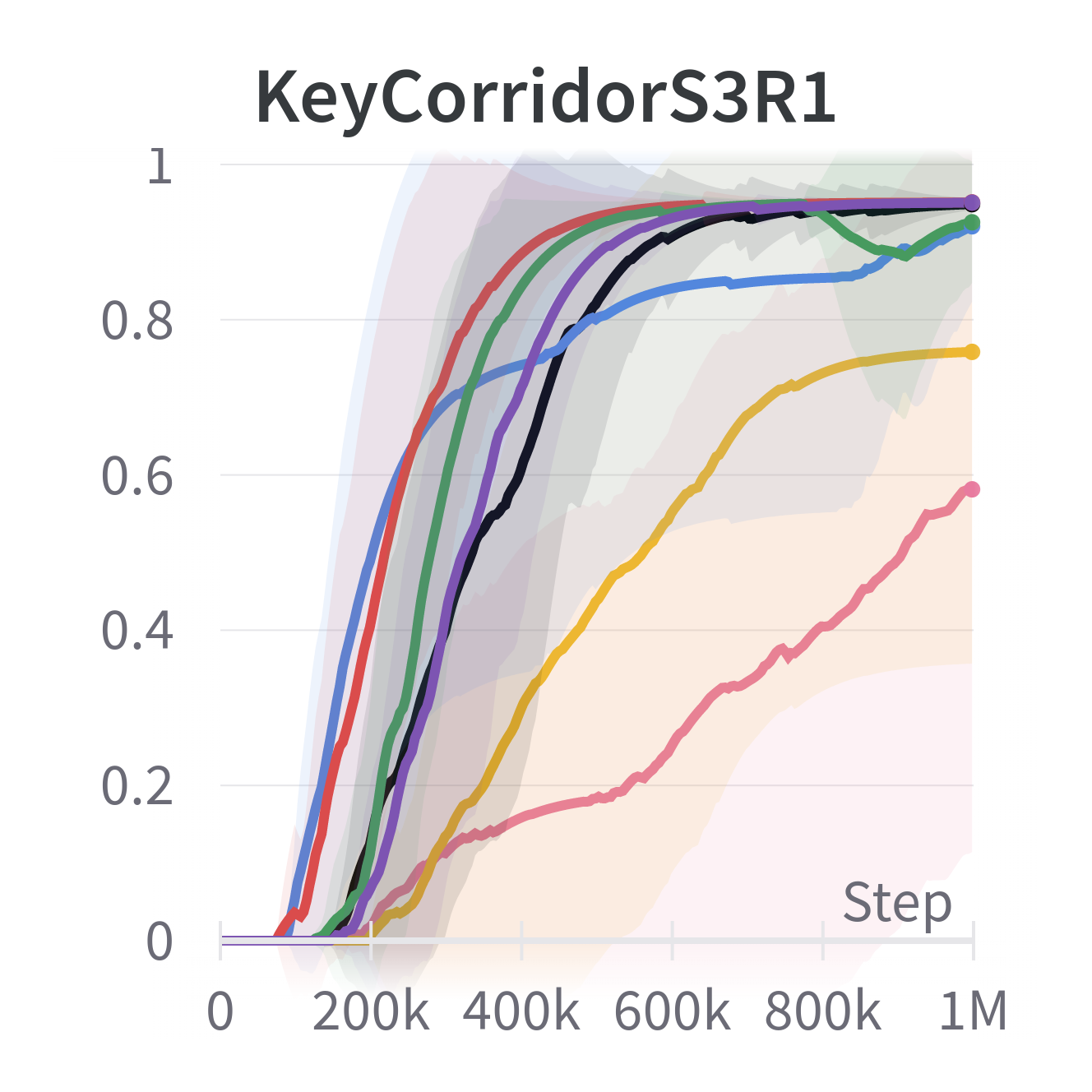} & 
      \includegraphics[width=0.23\textwidth]{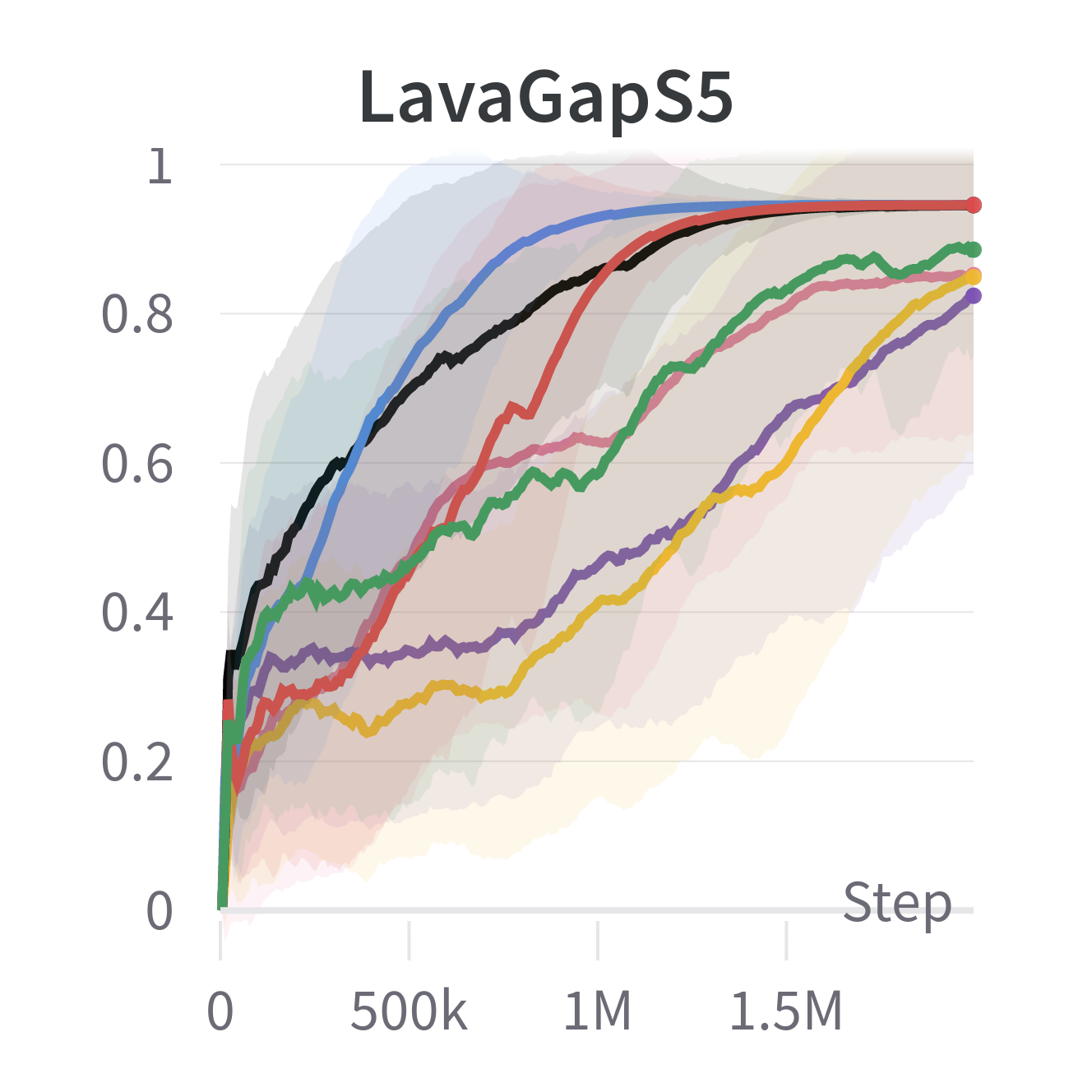} &
      \includegraphics[width=0.23\textwidth]{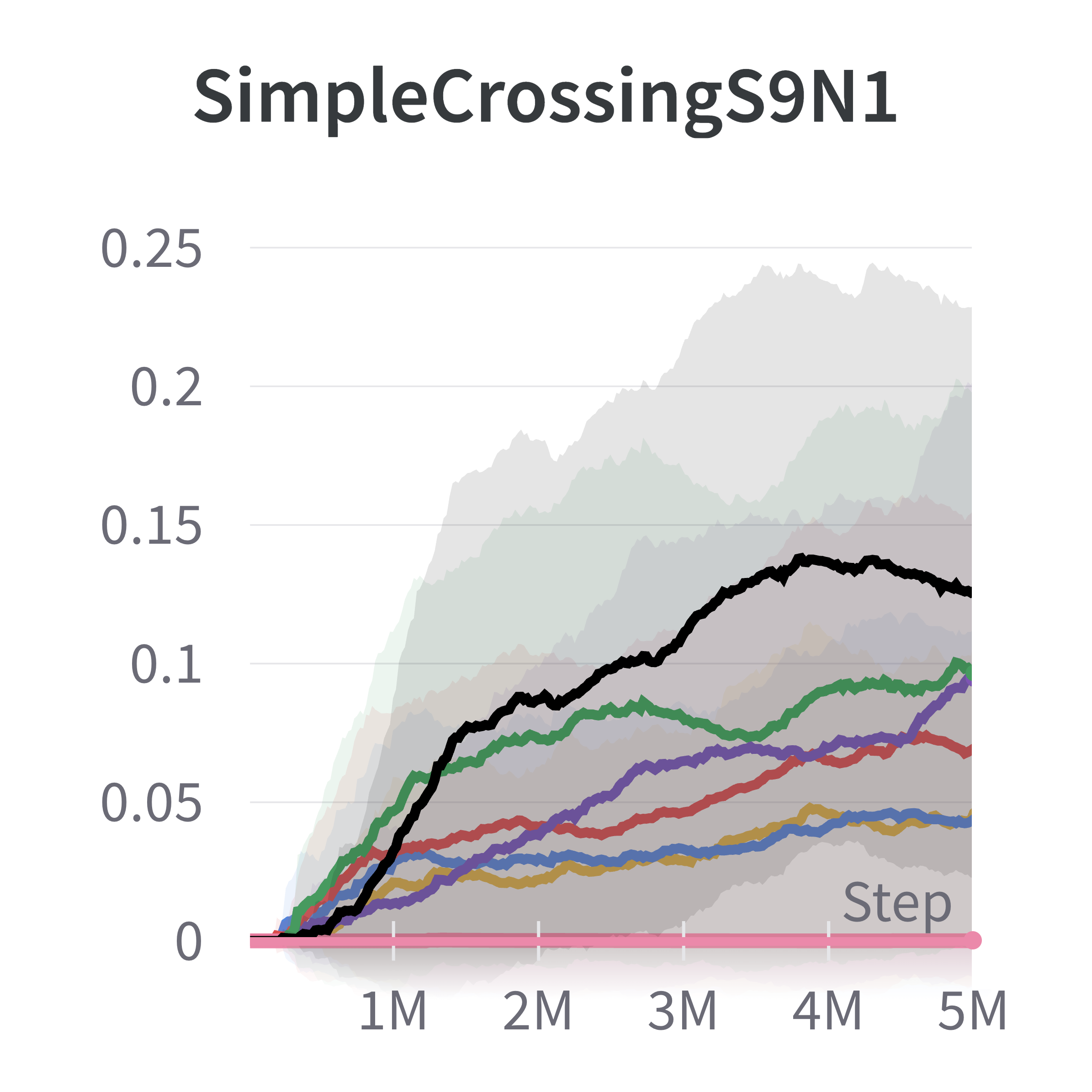} & 
      \includegraphics[width=0.23\textwidth]{figure/long-lavacrossings9n1.png} \\
      \includegraphics[width=0.23\textwidth]{figure/long-lavacrossings13n1.png} & 
      \includegraphics[width=0.23\textwidth]{figure/putnear-6x6-n2.png} & \includegraphics[width=0.23\textwidth]{figure/long-fetch-8x8-n3.png} & 
      \includegraphics[width=0.23\textwidth]{figure/long-multiroom-n2-s4.png} \\
      \includegraphics[width=0.23\textwidth]{figure/dynamic-obstacles-8x8.png} &
      \includegraphics[width=0.23\textwidth]{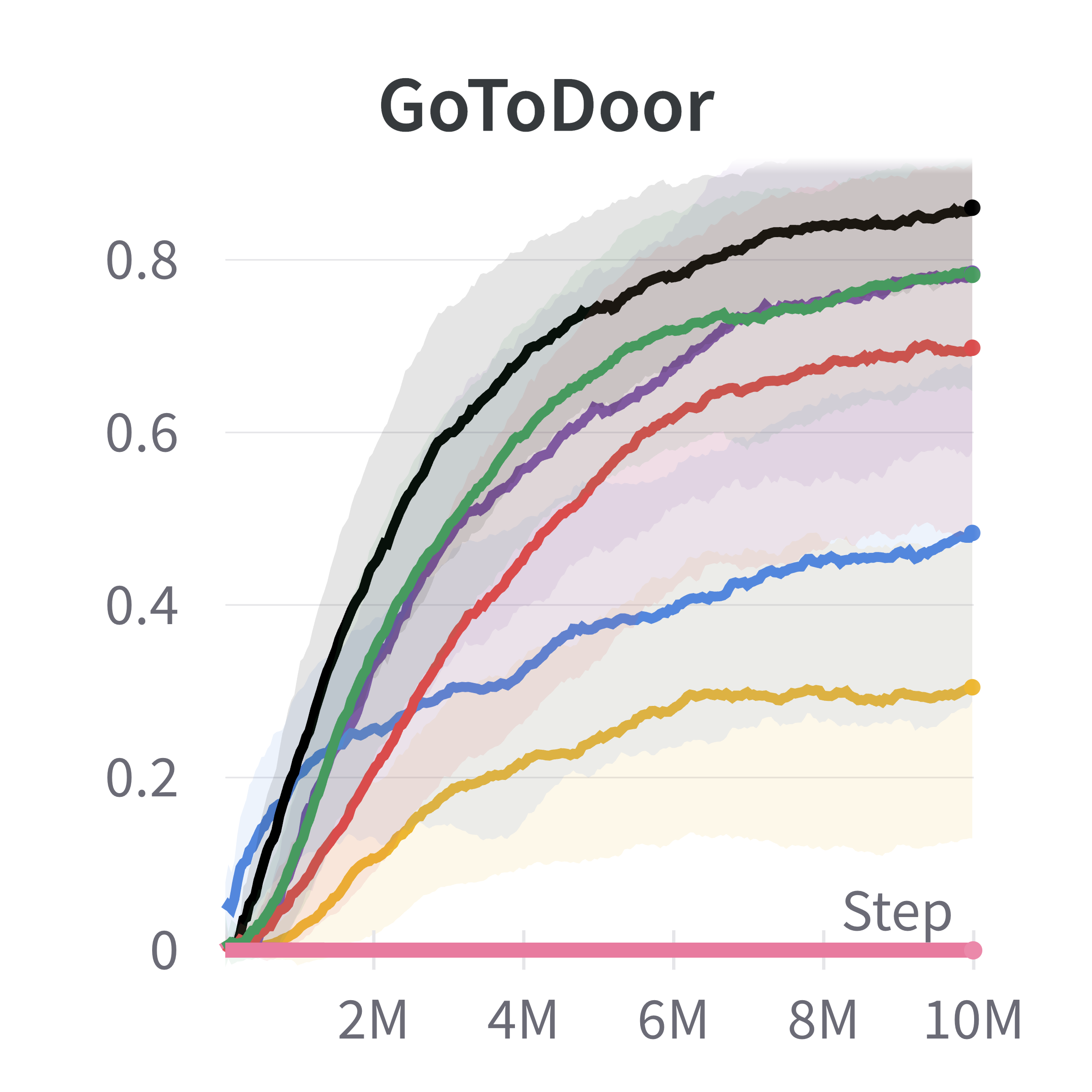} \\
\end{tabular}
\end{center}
\caption{Performance comparison in the MiniGrid tasks}
\label{fig:MiniGridOverallResults}
\end{figure*}

\begin{figure*}[hbt!]
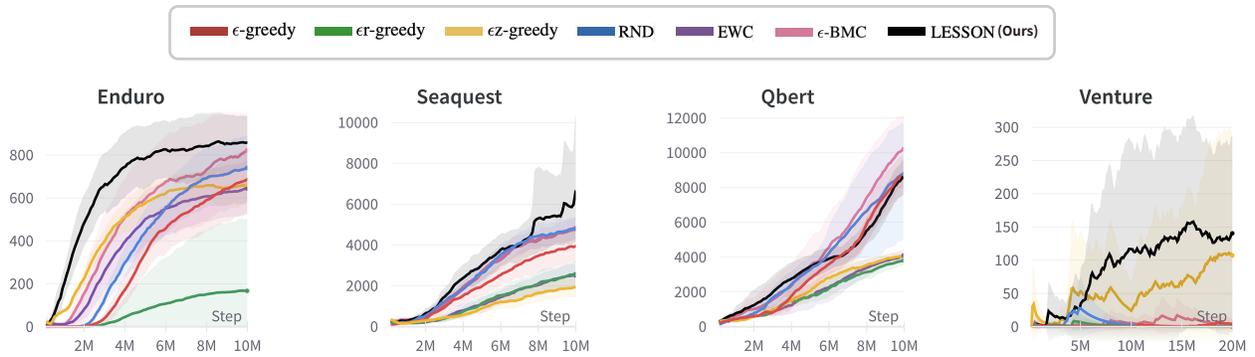

\begin{center}
\includegraphics[width=0.7\textwidth]{figure/label/label.pdf}
\begin{tabular}{cccc}
      \includegraphics[width=0.23\textwidth]{figure/enduro.png} &       \includegraphics[width=0.23\textwidth]{figure/seaquest.png} &
      \includegraphics[width=0.23\textwidth]{figure/qbert.png} &   \includegraphics[width=0.23\textwidth]{figure/venture.png} 
\end{tabular}
\end{center}
\caption{Performance comparison in the Atari tasks}
\label{fig:AtariOverallResults}
\end{figure*}

The results of fourteen MiniGrid environments and four Atari environments are provided in Fig. \ref{fig:MiniGridOverallResults} and Fig. \ref{fig:AtariOverallResults}, respectively.

\newpage

\newpage

\section{Experiment Result of Atari Montezuma's Revenge} \label{sec:appx-montezuma}

\subsection{Environment Specification}

Montezuma's Revenge  is considered as one of the most challenging exploration environments in the Atari game library. As shown in Fig. \ref{fig:montezuma} (a), the rooms where the player engages in gameplay contain various obstacles like ladders, ropes, platforms, enemies, and traps. To progress through the stages, the player must successfully navigate these obstacles by jumping, climbing, and employing precise timing. Additionally, the game encompasses multiple levels, each possessing its unique layout and distinctive difficulties. To increase their score and gain access to new regions, the player must explore the pyramid, discover concealed passages, keys, and other undisclosed secrets.

\subsection{Base model and Baselines}

Due to its difficulty, DQN based on a simple exploration method such as $\epsilon$-greedy takes too long time in learning  Montezuma's Revenge.  Therefore, we adopted the approach from DQN-PixelCNN \cite{ostrovski2017count}, which shows relatively faster learning in Montezuma's Revenge compared with other simple DQN variants. 

\subsubsection{DQN-PixelCNN}

DQN-PixelCNN is an advanced DQN variant  introduced by \citet{ostrovski2017count}. This method exploits a density model  to enable count-based exploration. 
The authors employed PixelCNN, an advanced neural density model designed for image data, as their neural density model, and efficiently computed pseudo-count, which  measures the novelty or unfamiliarity of states to promote exploration  in count-based exploration models. Specifically, by employing PixelCNN as the density model, the pseudo-count is computed  with the equation $ \hat{N}(x) = \rho (x)\hat{n}(x) $, where $\hat{n}(x)$ is a cumulative pseudo-count derived from the PixelCNN model's updated probability estimation $\rho'(x)$ \citep{ostrovski2017count}. This probability is computed immediately after training on the input sample $x$. Then, an intrinsic reward is determined based on the computed pseudo-count. Then, DQN-PixelCNN performs Q-learning with the weighted sum of extrinsic and intrinsic rewards.

\subsubsection{Modified Design of Intra-policies of LESSON}

Due to its effectiveness of the PixelCNN-based intrinsic reward generation, we removed 
the prediction-error maximizing (PEM) intra-policy based on RND \citep{burda2018exploration}, but included the pseudo-count maximizing (PCM) intra-policy based on PixelCNN instead. The PCM intra-policy selects the action
that maximizes the sum of pseudo-count, where the pseudo-count is computed based on PixelCNN of \citet{ostrovski2017count}.
The other intra-policies, i.e., greedy, random and TE-random intra-policies, remain as before.

\subsubsection{Modified Baselines}

In Section \ref{sec:experiments} of the main paper, LESSON was evaluated against the following six baselines: 

- $\epsilon$-greedy (vanilla DQN), 

- two simple DQN variants:  $\epsilon$z-greedy, $\epsilon$r-greedy 

- $\epsilon$-BMC, which learns $\epsilon$, 

- RND-based DQN,  

- and equal weight combining (EWC).

With the use of  DQN-PixelCNN, the following modifications were made: 

- The RND-based DQN baseline was replaced by DQN-PixelCNN.

- Consequently, EWC now randomly choose one out of $\epsilon$-greedy, $\epsilon$r-greedy, $\epsilon$z-greedy,  and DQN-PixelCNN  with equal probability.  Here, $\epsilon$-BMC was excluded for EWC because $\epsilon$ in $\epsilon$-BMC changes over time.

\subsection{Implementations}

DQN-PixelCNN and other baselines were implemented based on the code provided from \url{https://github.com/NoListen/ERL}.

\subsection{Performance}

\begin{figure}[hbt!]
\centering
\begin{tabular}{ccc}
      \raisebox{0.36\height}{\includegraphics[width=0.15\textwidth]{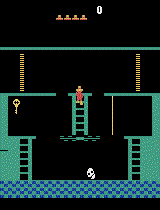}} \hspace{-2ex}
      & \raisebox{.0\height}{\includegraphics[width=0.52\textwidth]{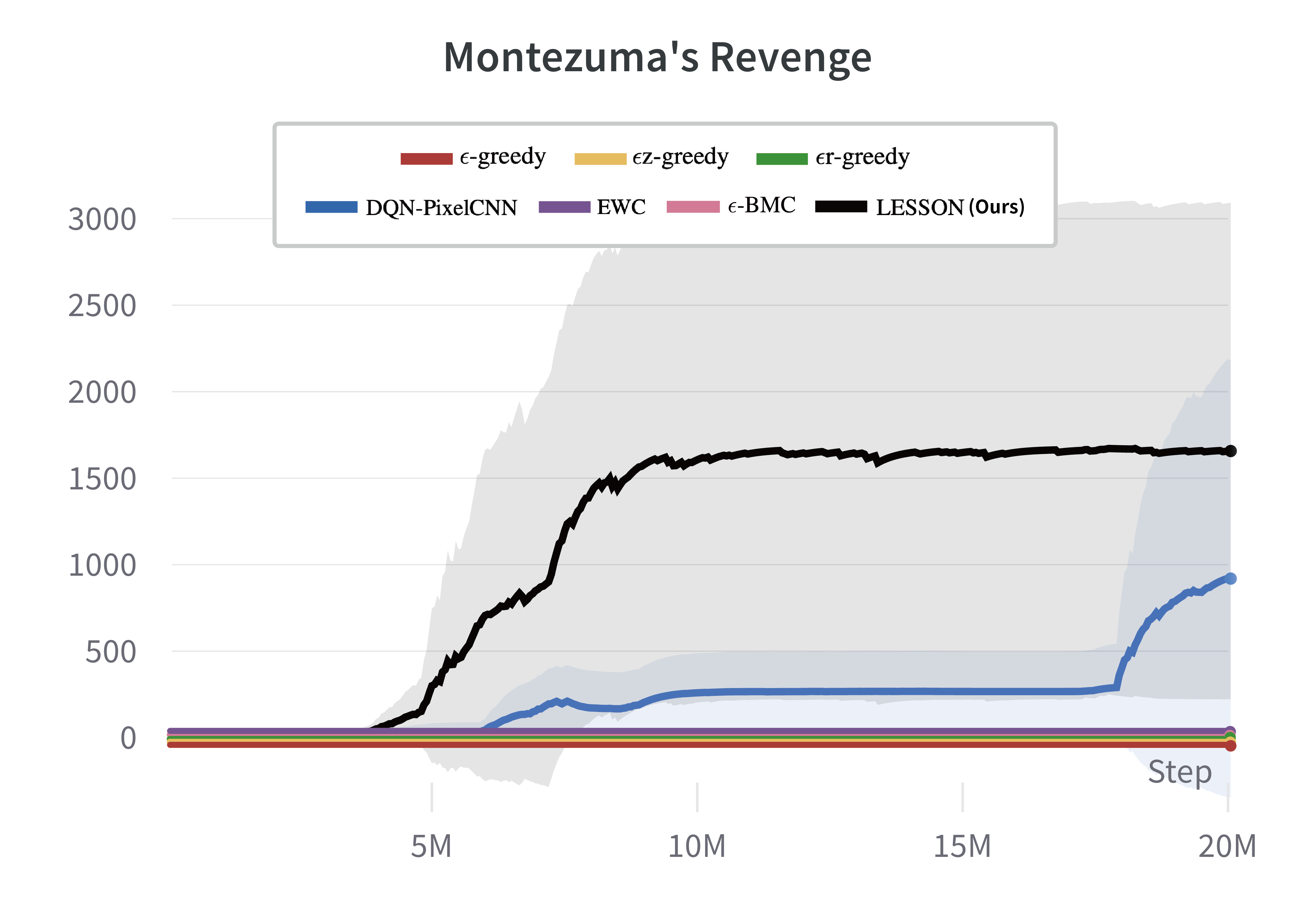}}
      & \raisebox{0.1\height}
      {\includegraphics[width=0.25\textwidth]{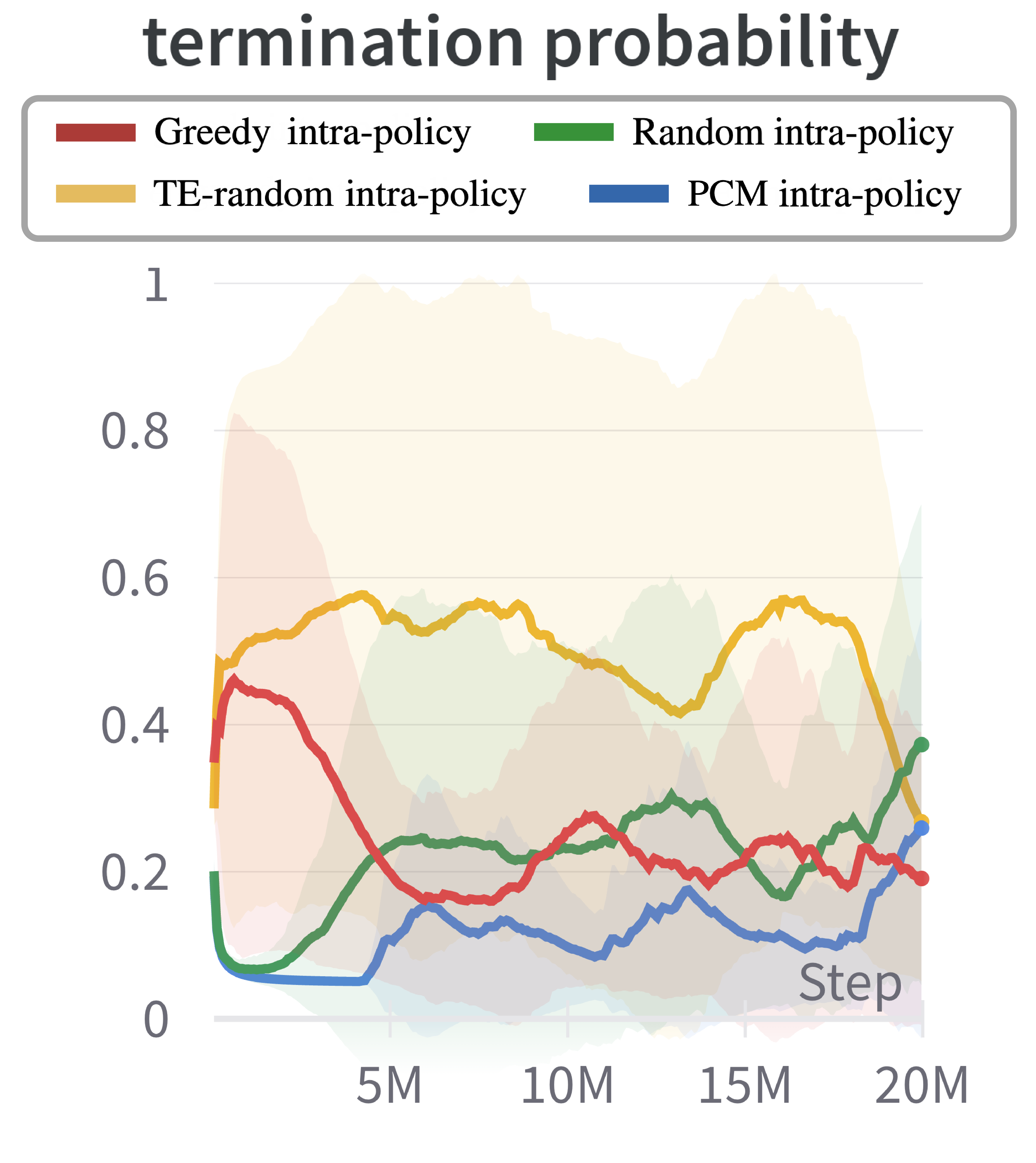}} \\
     (a) & (b) & (c) \\
    \end{tabular}
\caption{Performance on  Montezuma's Revenge. (a) the view of environment and (b) performance comparison (c) termination probability during training. The source of Fig. 14 (a) is \url{https://www.gymlibrary.dev/environments/atari}}.
\label{fig:montezuma}
\end{figure}

We evaluated LESSON and the baselines with three different seeds, and compared their performance. The result is shown in Fig. \ref{fig:montezuma} (b).  

The experimental result demonstrates that LESSON still shows significantly better performance compared with other baselines. 
Especially, LESSON including an PixelCNN-count-based exploration strategy as its intra-policy significantly outperforms DQN-PixelCNN itself, especially in terms of the learning speed. 
This result strongly suggests that {\em the mixture of randomness-based exploration  and history-based exploration is very effective for overall exploration in  difficult tasks such as  Montezuma's revenge.} LESSON provides a good framework to mix these exploration strategies.  Through this experiment, we  validated the effectiveness of LESSON in hard exploration tasks such as Montezuma's Revenge. 

Note that the point of LESSON employing DQN-PixelCNN is not comparable to that of 7570 after 1.6 billion frames reported in  \citep{burda2018exploration}. This seems to be a consequence  from the difference in the target policy.  We used an 1-step TD Q-learning baseline for our target policy, whereas \citet{burda2018exploration} used PPO which exploits $n$-step advantage estimation. 

Since LESSON is a general framework to integrate multiple distinct exploration strategies together with  the greedy target policy based on option-critic, it can be applied to  other state-of-the-art algorithms designed for 
target policy, which remains as a future work. We expect that applying LESSON to other state-of-the-art algorithms beyond DQN can  enhance their performance further.

\newpage
\section{In-Depth Analysis of Exploratory Behaviour}

\begin{figure}[h]
    \centering
    \begin{tabular}{ccc}
      \raisebox{0.2\height}{\includegraphics[width=0.20\textwidth]{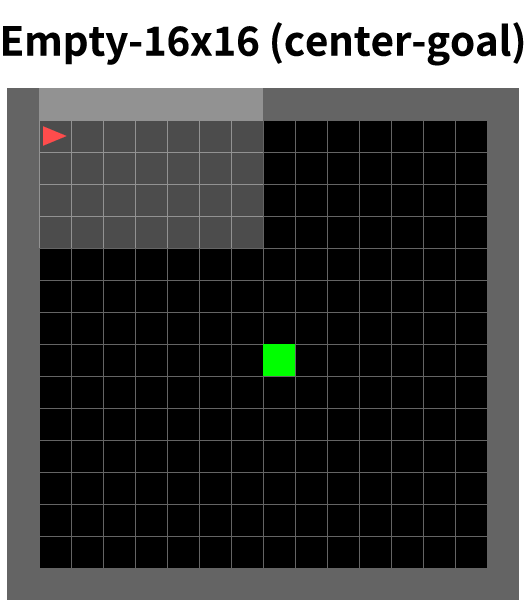}} &
      \raisebox{.0\height}{\includegraphics[width=0.27\textwidth]{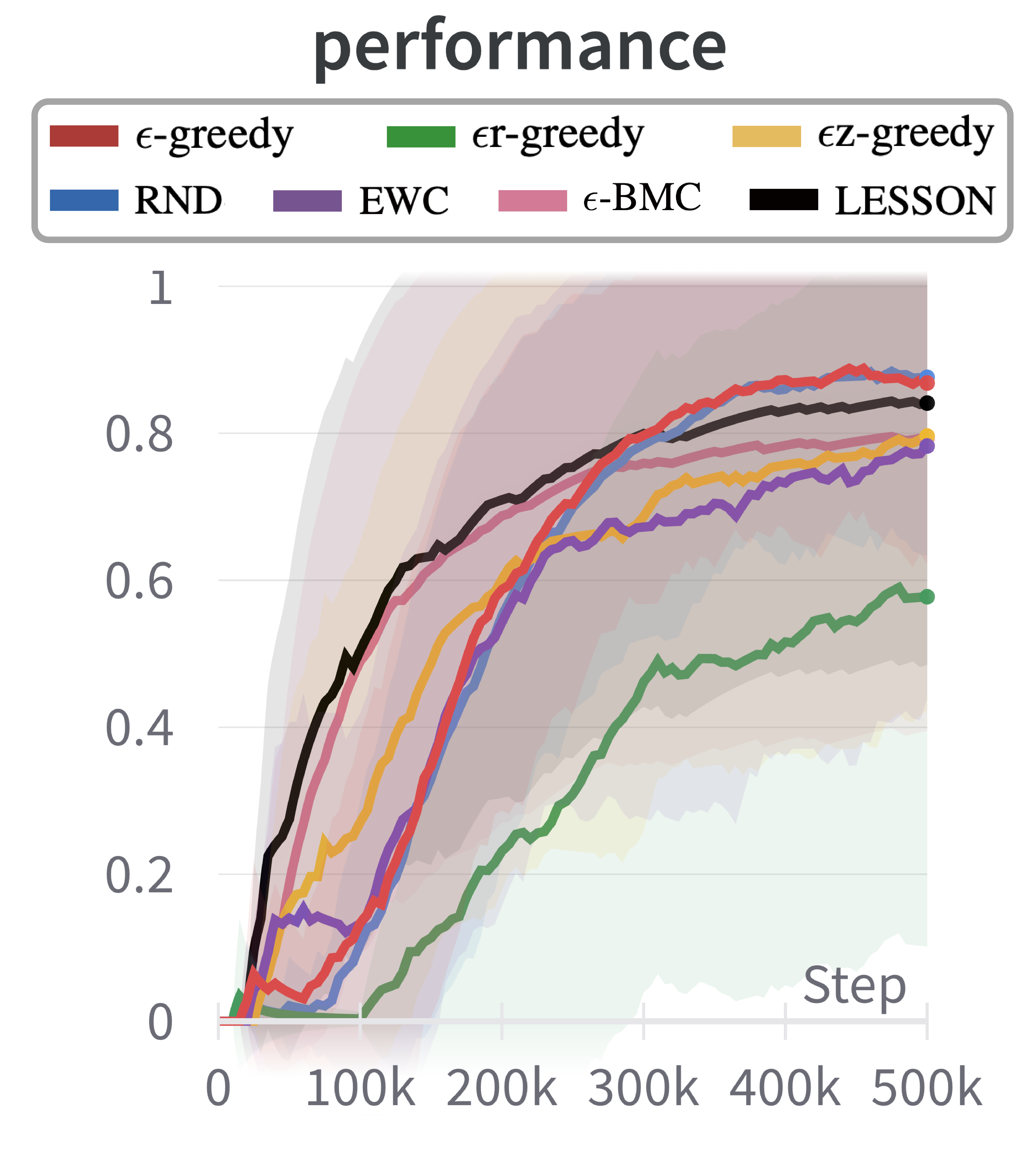}} & 
      \raisebox{.0\height}
      {\includegraphics[width=0.27\textwidth]{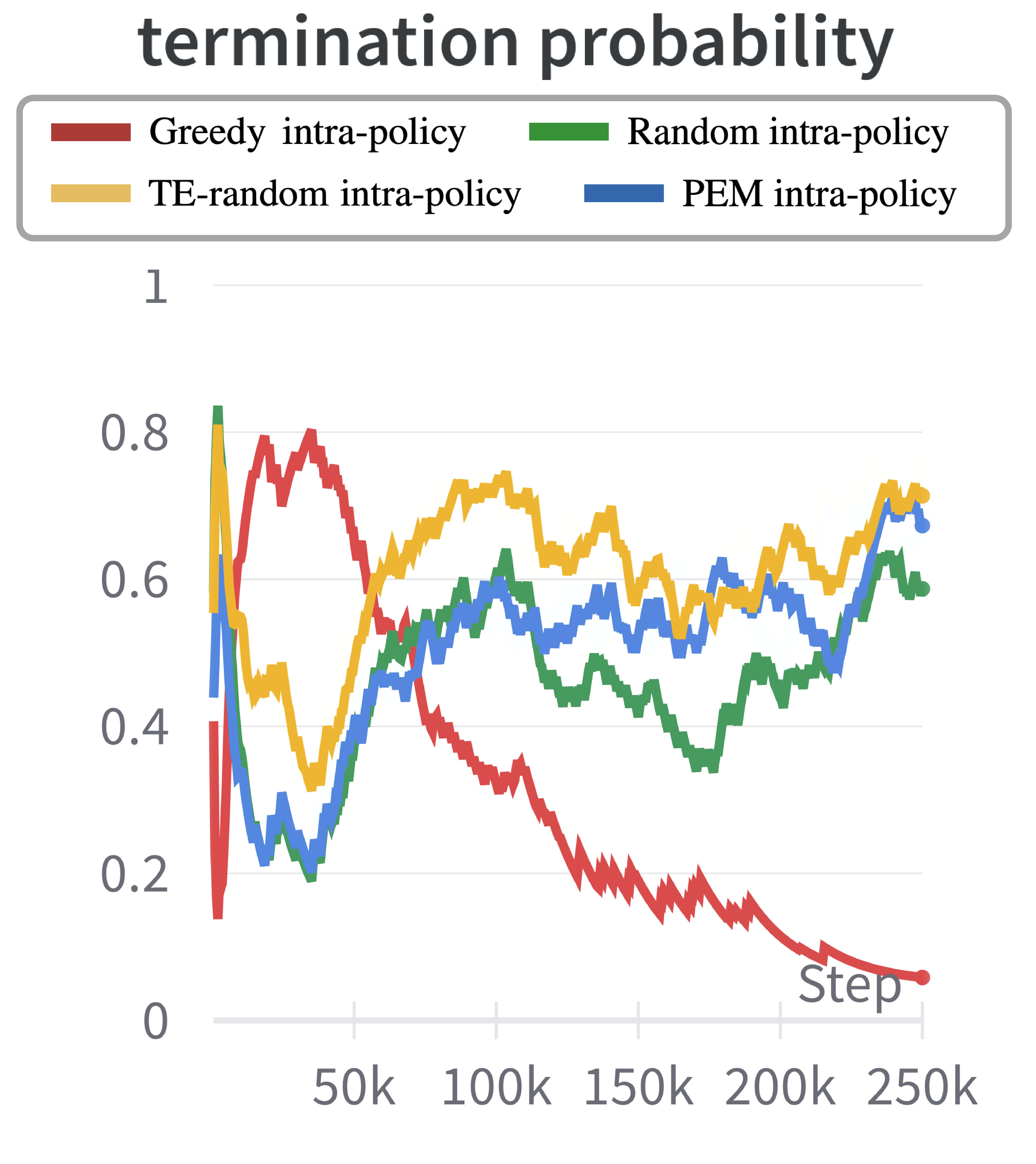}} \\
     (a) & (b) & (c)  \\
    \end{tabular}
        \begin{tabular}{c}
       \includegraphics[width=0.75\textwidth]{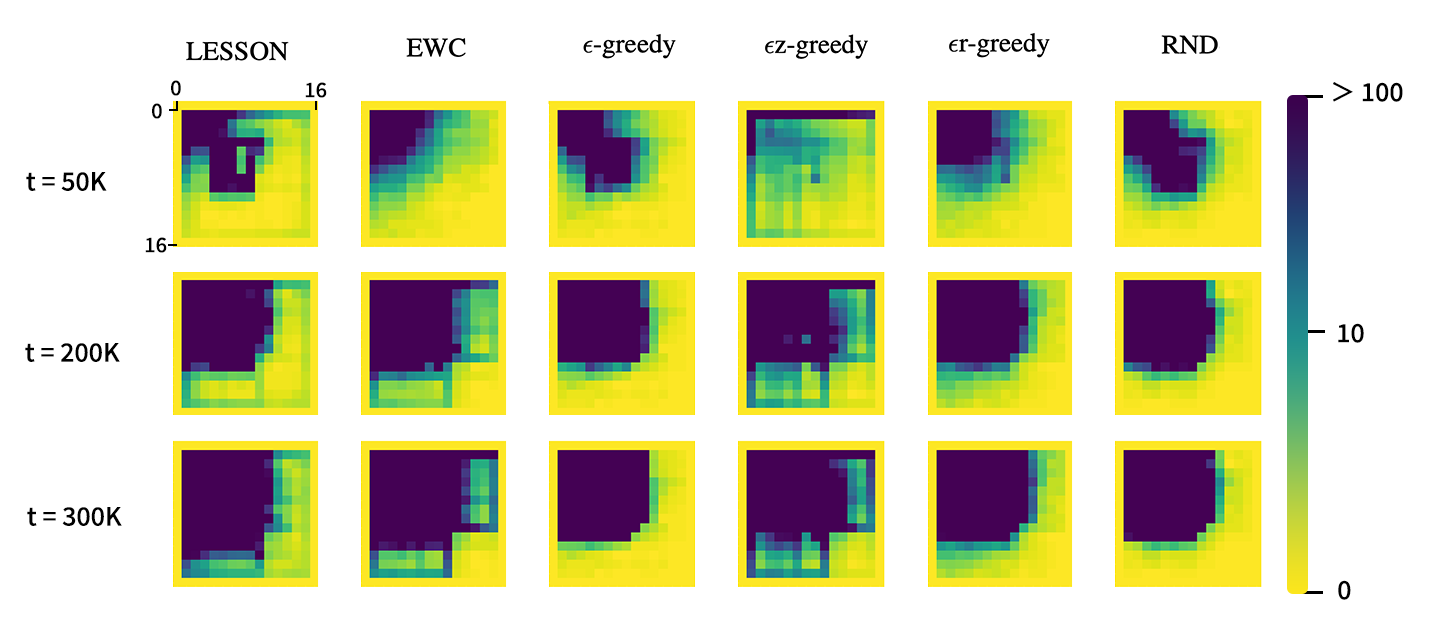}\\
    (d) \\
    \end{tabular}
\caption{Similarly to Figure \ref{fig:visitation}, a comparison between LESSON and the baseline is presented. In this case, the task of reaching the goal is performed on the same empty 16x16 grid, but the goal position has been shifted to the center: (a) the view of environment from the MiniGrid Empty-16x16 environment, (b) performance comparison, (c) termination probabilities of the intra-policies of LESSON during training, and  (d) visualization of visited states. }
\label{fig:visitationappend}
\end{figure}

In addition to the exploratory behavior analysis  discussed in Section \ref{sec:analysis}, we conducted another Empty-16x16 environment where the green goal is placed at the center of the map. The result is shown in Fig. \ref{fig:visitationappend}. As seen in  Fig. \ref{fig:visitationappend} (c),  the termination probability of the greedy policy in the center-goal environment decreases at a faster rate compared with  that of the bottom-goal environment shown in Fig. \ref{fig:visitation}. This result shows  that LESSON can automatically control the necessary level of exploration in contrast to  the conventional  $\epsilon$-greedy approach which requires manual adjustment of the $\epsilon$ parameter to control exploration.

\end{document}